\documentclass[runningheads]{llncs}

\usepackage[mobile]{eccv}

\usepackage{multirow}
\usepackage{colortbl} 

\usepackage{eccvabbrv}
\usepackage{pifont} 

\usepackage{graphicx}
\usepackage{booktabs}

\usepackage{microtype}
\usepackage{booktabs} 

\usepackage{algorithm2e}

\usepackage{url}

\usepackage[skip=4pt]{caption}
\usepackage{pythonhighlight}
\usepackage{listings}
\usepackage{inconsolata}

\definecolor{citecolor}{HTML}{0071BC}
\definecolor{linkcolor}{HTML}{ED1C24}

\usepackage{amsmath}
\usepackage{amssymb}
\usepackage{mathtools}

\usepackage[textsize=tiny]{todonotes}

\usepackage[accsupp]{axessibility}  

\newcommand{\cmark}{\ding{51}}
\newcommand{\xmark}{\ding{55}}

\usepackage[pagebackref,breaklinks,colorlinks,citecolor=eccvblue]{hyperref}

\usepackage{orcidlink}
\usepackage{soul}

\begin{document}

\title{PaPr: Training-Free One-Step Patch Pruning with Lightweight ConvNets for Faster Inference} 


\titlerunning{PaPr for Faster Inference}


\author{Tanvir Mahmud$^{1}$ \ Burhaneddin Yaman$^{2}$ \ Chun-Hao Liu$^{3}$ \ Diana Marculescu$^{1}$ \\
}

\authorrunning{T. Mahmud et al.}

\institute{University of Texas at Austin\\
\email{\{tanvirmahmud, dianam\}@utexas.edu}
\and
Bosch Research North America\\
\email{burhaneddin.yaman@us.bosch.com}
\and
Amazon Prime Video\\
\email{chunhaol@amazon.com}\\
}
\maketitle

\begin{abstract}

As deep neural networks evolve from convolutional neural networks (ConvNets) to advanced vision transformers (ViTs), there is an increased need to eliminate redundant data for faster processing without compromising accuracy. Previous methods are often architecture-specific or necessitate re-training, restricting their applicability with frequent model updates. To solve this, we first introduce a novel property of lightweight ConvNets: their ability to identify key discriminative patch regions in images, irrespective of model's final accuracy or size. We demonstrate that fully-connected layers are the primary bottleneck for ConvNets performance, and their suppression with simple weight recalibration markedly enhances discriminative patch localization performance. Using this insight, we introduce PaPr, a method for substantially pruning redundant patches with minimal accuracy loss using lightweight ConvNets across a variety of deep learning architectures, including ViTs, ConvNets, and hybrid transformers, without any re-training.  Moreover, the simple early-stage one-step patch pruning with PaPr enhances existing patch reduction methods. Through extensive testing on diverse architectures, PaPr achieves significantly higher accuracy over state-of-the-art patch reduction methods with similar FLOP count reduction. More specifically, PaPr reduces about 70\% of redundant patches in videos with less than 0.8\% drop in accuracy, and up to $3.7\times$ FLOPs reduction, which is a 15\% more reduction with 2.5\% higher accuracy. Code is released at \url{https://github.com/tanvir-utexas/PaPr}.
\end{abstract}
\section{Introduction}
\label{sec:1_intro}

Deep neural networks have grown from simple convolutional neural networks (ConvNets) to complex transformer models, aiming for better accuracy with more computations~\cite{resnet, vit}. Vision transformers (ViTs) excel by focusing on important parts of images with long-range attention and large-scale pre-training~\cite{mae, clip}. This success, however, comes at the cost of increased computational cost. Operating these large networks efficiently for downstream applications is crucial. Most visual tasks demand precisely pinpointing key image regions against complex backgrounds. Higher image resolution improves accuracy by capturing more details but also adds unnecessary background processing, burdening large models~\cite{swag}. This highlights the need to cut down on redundant data in high-resolution images to maintain both speed and performance with advanced techniques~\cite{tome, adavit}.

\begin{figure}[t]
\centering
\includegraphics[width= 1.0\textwidth]{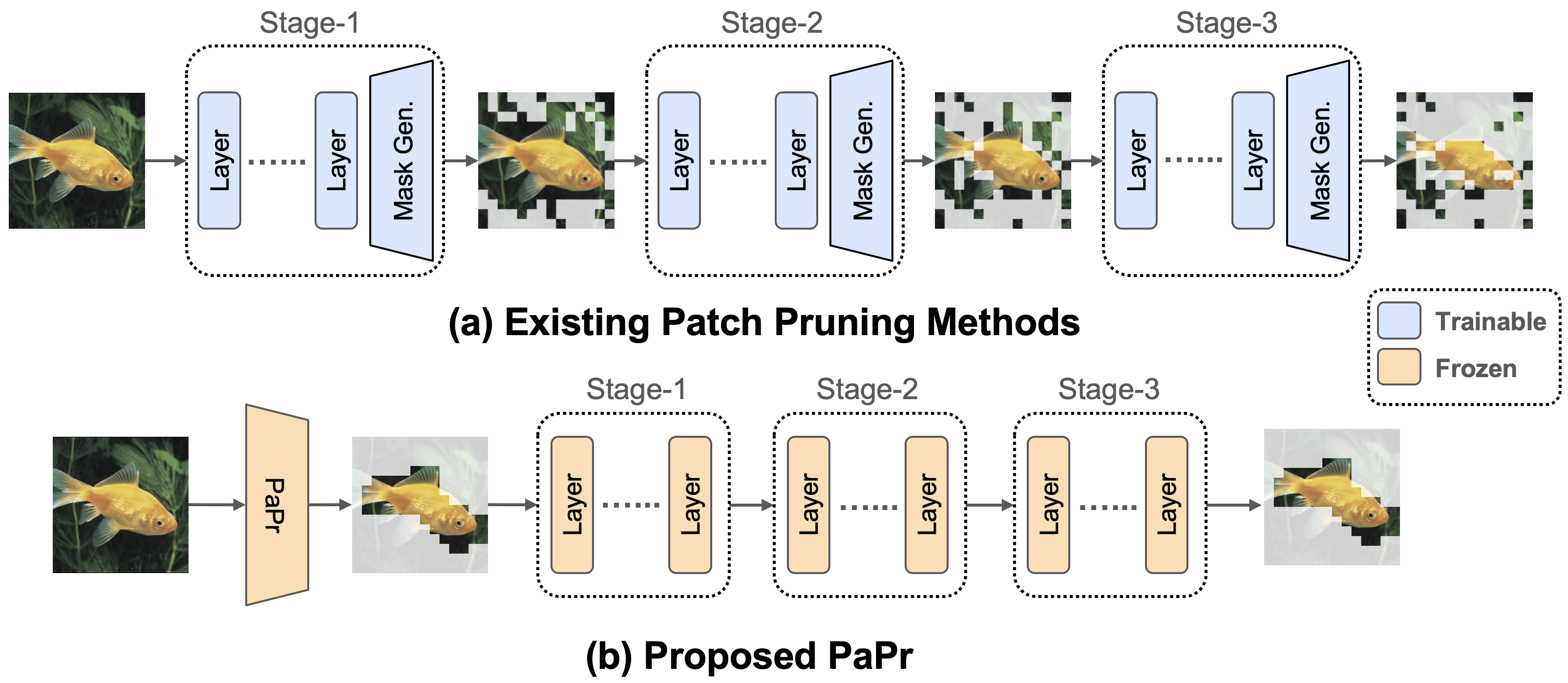}
\caption{(a) Existing patch pruning methods gradually reduce patches over the model. This requires additional training of mask generators in intermediate layers. (b) Proposed PaPr directly prunes redundant patches early in the network by leveraging pretrained lightweight ConvNets  and directly speeds-up \textit{off-the-shelf} models without re-training.}
\label{fig:1}
\vspace{-4mm}
\end{figure}

Identifying key image regions demands a comprehensive understanding of the image and model operations. Incorrect key region estimation can impair pre-training performance by eliminating crucial areas. Traditional approaches~\cite{Avit, spvit, evit, tome, adavit} for pruning redundant patch regions are hindered by three main limitations: (1) They often necessitate complex training of extra modules, which becomes increasingly difficult as baseline models evolve with more data, enhanced training methods, and deeper structures~\cite{dynamicvit, Avit, spvit}. Re-training these modules for each model update is impractical. (2) Without a complete image understanding, these methods incrementally prune patches across the network, leading to unnecessary computations in early layers—particularly problematic for deeper models. (3) Many rely on specific architectural features for patch reduction, such as class tokens or attention maps, limiting their use to a narrow set of network designs~\cite{evit, ats}. Hence, there is a pressing need for a patch pruning solution that is adaptable to various modern architectures without additional training, can eliminate redundant patches in a single-step, thereby making it suitable for a broad spectrum of networks while streamlining each model update  (See Fig. \ref{fig:1}).

Recent work~\cite{tome, xu2024gtp, tokenfusion} focuses mostly on transformer based architectures rather than ConvNets for patch reduction due to their impressive performance on various tasks, but they suffer from the aforementioned limitations. 
While ConvNets achieve lower ImageNet-1k top-$1$ accuracy than large ViTs (68.7\% in MobileOne-S0~\cite{mobileone} \textit{vs.} 88.7\% in ViT-Huge~\cite{swag}), they exhibit a remarkable ability to efficiently process the key image regions with hierarchical inductive bias. 
Our empirical investigation reveals that, as we broaden the evaluation metric (increasing $k$ in top-$k$ evaluations, see Fig. \ref{fig:4}), the benefits of deeper models diminish, especially with a large number of image classes (\textit{e.g.}, 1000 in ImageNet). This suggests that shallower models excel at identifying discriminative areas as their bigger counterparts, rendering them ideal for patch pruning.

Leveraging this insight, we propose PaPr, a novel Patch Pruning method that employs pretrained lightweight ConvNets for efficient, one-step patch pruning in a wide variety of deep learning models, maintaining accuracy while significantly cutting computational demands. 
Our findings show lightweight ConvNets  have remarkable ability in identifying discriminative image regions but struggle in fine-grained prediction. To address this, PaPr relies on Patch Significance Maps (PSMs), which are generated using only the convolutional layers of ConvNets through uniform class weight recalibration in FC layers. Astonishingly, PSMs consistently highlight critical image regions across ConvNets of varying sizes and accuracies (Fig. \ref{fig:6}), thereby amplifying the efficacy of ultra lightweight ConvNets.

ConvNets inherently preserve the positional property of patches for their inductive bias, while ViTs use cross-attention to blend patch features, leading to variability in patch relevance across models. Unlike previous methods that prune redundant patches gradually over multiple steps across intermediate layers of ViTs~\cite{tome, dynamicvit, tokenfusion}, our approach, PaPr, simplifies this process by eliminating non-essential patches at once, immediately after extraction {(see Fig. \ref{fig:3})}, by leveraging lightweight ConvNets to assess patch significance. This direct, one-step pruning approach significantly cuts computational demands and is also compatible with other patch reduction techniques (See Fig.~\ref{fig:5}). PaPr's ability to separate crucial patch identification from fine-grained class prediction enhances a wide range of pre-trained models without further training, ensuring high accuracy with notable speed ups. By bypassing the complex training required for conventional patch selectors, PaPr capitalizes on the comprehensive capabilities of minimalist ConvNets for efficient patch pruning in larger models.

Our experiments demonstrate PaPr's effectiveness across various architectures and pre-training methods, achieving significant reduction in redundant patches for ViTs, large-scale ConvNets (\textit{e.g.}, ConvNext~\cite{convnet2020}), and hybrid transformers (\textit{e.g.}, Swin~\cite{swin}), outperforming state-of-the-art (SOTA) patch reduction methods by a large margin. Notably, PaPr can be easily integrated with most existing patch reduction methods to reduce patches early in operation. More specifically, PaPr boosts ToMe~\cite{tome} accuracy by 4.5\% for a similar computational budget with ViT-B.
Moreover, PaPr can accelerate training akin to token merging techniques, unlike most patch pruning methods that fail to boost training speed~\cite{dynamicvit, adavit, Avit}.
PaPr shows robust patch localization performance with ultra lightweight ConvNets (<0.3\% accuracy loss) for 42$\times$ reduction in proposal FLOPs, thereby enabling its use for larger \textit{off-the-shelf} models.
Remarkably, in video recognition, PaPr cuts down around 70\% of redundant patches, resulting in up to 3.7$\times$ FLOPs reduction with minimal impact on accuracy ($\approx0.8\%$). 
In addition, extensive qualitative visualizations  demonstrate the effectiveness of PaPr over existing methods.

We summarize our main contributions as follows:
\begin{itemize}
    \item We propose PaPr, a novel background patch pruning method that can seamlessly operate with ViTs, ConvNets, and hybrid transformers, without further training while leveraging batch processing.

    \item We propose a simple weight recalibration method in ConvNets to precisely and efficiently locate discriminative patches, irrespective of model size.
    
    \item We facilitate the use of ultra-lightweight ConvNets to speed-up large models, such as ViTs, with a seamless framework and negligible accuracy loss.
    
    \item We present extensive qualitative and quantitative results across numerous model architectures in both image and video applications. 
\end{itemize}
\vspace{-0.2cm}

\vspace{1em}
\vspace{-5mm}
\section{Related Work}
\vspace{-1mm}
\label{sec:2_related}

\subsection{From ConvNets to Vision Transformers}
\vspace{-1mm}

ConvNets have been pivotal in computer vision, offering computational efficiency through kernel reuse and localization\cite{resnet, efficientnet, shufflenet, repvgg, mobilenets, mobilenetv2, mobilenetv3, mcunet, mobileone}. Yet, they struggle in capturing long-range dependencies, a gap bridged by ViT~\cite{vit}. Inspired by the success of transformers from natural language processing (NLP), ViTs excel in long-range feature modeling through cross-attention, outperforming ConvNets albeit with higher computational demands and extensive training data requirements~\cite{wang2020linformer,chu2021twins, castingvit, deit}. Efforts to alleviate these issues include self-supervised\cite{mae, maskedvit, sparsemae, videomae} and weakly-supervised~\cite{swag, clip} pre-training, although challenges in computational complexity and optimization remain. Hybrid architectures, merging ConvNets' inductive biases with ViTs' cross-attention, offer a balanced solution by reducing computational load while maintaining performance~\cite{hierarchyViT, wu2021cvt, mobilevit, swin, swinv2, cswin, mvit, mvit2, levit}. Recent ConvNets advancements~\cite{convnet2020, convnextv2}, with improved training and large-scale data, now rival ViTs, questioning if architectural innovations or enhanced training primarily drive performance gains.

Addressing the computational demands of these models, we focus on enhancing operational efficiency by pruning redundant patches without architectural modifications or re-training, maintaining performance while streamlining updates.

\vspace{-2mm}
\subsection{Class Activation Mapping for Explainable Deep Learning}
\vspace{-1mm}

Class Activation Mapping (CAM) techniques provide explainable visual reasoning for neural network predictions by highlighting activation regions\footnote{``Patch'' and ``region'' have been used interchangeably based on the context.} crucial for decisions~\cite{basecam, fcam, ucam, gradcam, bettercam}. While ConvNets utilize convolutional operations to preserve spatial information, making them apt for such visualizations, ViTs link attention weights to class tokens for prediction~\cite{dino, dinov2}. However, CAM's reliance on accurate class predictions for effective feature localization is a significant drawback, as ConvNets' lower accuracy compared to ViTs may lead to incorrect object localization. Despite ViTs' superior fine-grained classification capabilities, it's unclear if they prioritize the same discriminative regions as ConvNets. Additionally, CAM approaches often require gradient tracking~\cite{gradcam, gradcam++} or complex feature map decomposition~\cite{eigencam}, complicating batch processing and necessitating extra optimization steps.

Our work diverges from traditional CAM by aiming to consistently identify discriminative patches across different architectures, focusing computational resources on the most relevant areas without being constrained by class prediction accuracy. This approach enables more efficient processing by allowing larger models to concentrate on the most significant regions, identified by lighter networks, thus significantly reducing computational demands.

\subsection{Patch Reduction for Faster Inference}

Several approaches have sought to enhance computational efficiency by reducing redundant patches in neural networks, with early strategies involving additional adapters or controllers to identify and prune less significant patches~\cite{dynamicvit,  unifiedpruning, learnedtokenpruning, colbertpruningstudy}. These methods, however, necessitate separate adapter training for each network and are slow to adapt due to the need to learn from the dynamics of other layers. In the context of ViTs, efforts have leveraged architectural features, such as class tokens and attention maps, for patch relevance, yet these solutions often fail to generalize across various architectures like hybrid transformers~\cite{swin} or larger ConvNets~\cite{convnet2020, convnextv2}, thus limiting their applicability~\cite{ats, spvit, Avit, adavit}. Furthermore, while some have explored patch merging in ViTs and a combination of merging and pruning~\cite{tokenfusion, pumer, tome}, these approaches lack a holistic image understanding by not considering the entire image context in their optimizations, leading to incremental and sub-optimal patch reduction.

Our work diverges from these traditional patch reduction methods by proposing a single-step, early-network patch removal strategy, that seamlessly integrates with any architecture without re-training.

\section{Methodology} 
\label{sec:approach}

\begin{figure}[t]
\centering
\includegraphics[width= 0.95\textwidth]{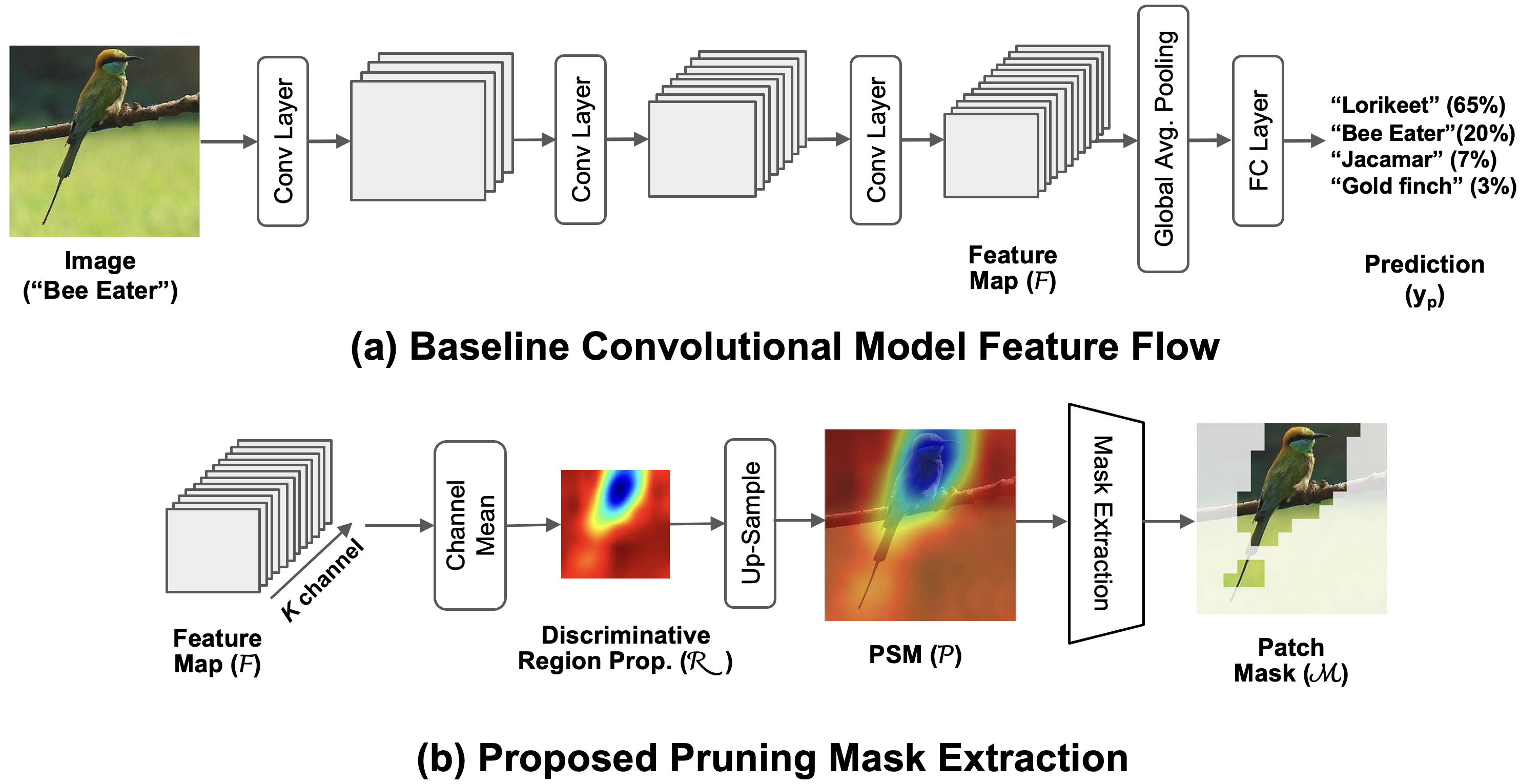}
\caption{(a) Baseline ConvNet gradually reduces the feature map to produce $\mathcal{F} = \{f_{k}(x, y) \}_{k=1}^K$, followed by global average pooling and fully connected (FC) layers to predict $y_p$. (b) In PaPr, we operate on $\mathcal{F}$ by suppressing the FC layer. Initially, we extract pixel mean over $K$ channels to produce discriminative region proposal $\mathcal{R}$. Later, simple upsampling operation generates the patch significance map (PSM) $\mathcal{P}$ of target dimension. Finally, patch mask $\mathcal{M}$ for top $z\%$ patches is obtained from $\mathcal{P}$.}
\label{fig:2}
\vspace{-2mm}
\end{figure}

Our methodology introduces a novel approach to discriminative patch pruning across various deep learning architectures, leveraging the innate capabilities of lightweight ConvNets. By generating a PSM, we efficiently identify and prune non-essential patches in a single step, enhancing computational efficiency without compromising accuracy. This process is universally applicable, seamlessly integrating with ViTs, ConvNets, and hybrid models, thereby addressing the limitations of previous methods with a scalable, architecture-agnostic solution.

\subsection{Extracting Discriminative Regions with ConvNets}
Despite achieving relatively lower top-1 accuracy on the ImageNet benchmark, lightweight ConvNets exhibit a competitive edge in top-10 accuracy when compared to their larger ViT counterparts (Fig.~\ref{fig:4}). This phenomenon underscores ConvNets' capability to effectively localize regions of interest through their convolutional layers, despite potential limitations in fine-grained classification attributed to the fully-connected (FC) layer. Our methodology leverages this insight by proposing discriminative patch regions while minimizing the influence of the FC layer (see Fig.~\ref{fig:2}), thereby enhancing the large \textit{off-the-shelf} model's focus on most salient image regions proposed by lightweight ConvNets (see Fig.~\ref{fig:3}).

Given an input image $X \in \mathbb{R}^{H \times W \times 3}$, with height $H$ and width $W$, we denote by $f_k(x, y)$ the feature map generated by the $k^{th}$ kernel in the last convolutional layer of typical ConvNets, $\forall k \in \{1, 2, \dots, K\}$. Here, each pixel $(x, y)$ in $f_k(x, y)$ corresponds to a patch window of size $(H/d\times W/d)$ in the input image $X$, reflecting a spatial down-scaling factor of $d$ through the convolutional layer stack. Typically, global average pooling (GAP) is applied to each feature map $f_k(x, y)$, yielding ${F} = \{F_k\}_{k=1}^K \in \mathbb{R}^{K}$, where ${F_k} = \sum\limits_{x, y} f_k(x, y) \in \mathbb{R}$. Finally, a fully connected (FC) layer ${W} \in \mathbb{R}^{C \times K}$ with elements $w_c^k$ processes ${F}$ to generate class predictions $y_p \in \mathbb{R}^{C}$ across $C$ classes as follows:
\begin{equation}
    y_p = WF = \sum\limits_{c} \sum\limits_{k} w_c^k \sum\limits_{x, y} f_k(x, y) = \sum\limits_{x, y} R(x, y),
\end{equation}
where $R \in \mathbb{R}^{h \times w}$ encapsulates the weighted mean class activation mapping of the convolutional feature map.

We note that, the reliance on class activation weights $w_c^k$ is influenced by the model's final accuracy, posing a challenge for smaller models. Our objective transcends mere accuracy enhancement, aiming to precisely locate discriminative patches rich in information irrespective of the model size or its final classification performance. Recognizing that discriminative region localization is pivotal for detailed classification, and acknowledging the competitive top-10 accuracy of lighter ConvNets, we posit that an extremely lightweight ConvNet suffices for initial discriminative region proposal.

To counteract the influence of the weak linear FC layer $W$ in convolutional region proposal, we propose an adjustment where $w_c^k = 1/KC, \ \forall k\in\{1, \dots, K\}$, facilitating the generation of a discriminative region proposal $\mathcal{R} \in \mathbb{R}^{h \times w}$, where
\begin{equation}
    \mathcal{R}(x, y) = \sum\limits_{c} \sum\limits_{k} \frac{1}{KC} f_k(x, y) = \frac{1}{K}\sum\limits_{k} f_k(x, y).
\end{equation}

This strategy allows us to leverage ConvNets for what they excel at: pinpointing critical image areas. By reducing reliance on class activation weights, we efficiently generate discriminative region proposals directly from convolutional outputs. This approach not only enhances the interpretability and efficiency of the localization process but also enables the application of more complex models for subsequent detailed analysis, optimizing the use of computational resources.

\subsection{Patch Significance Map}

With the discriminative region proposal $\mathcal{R} \in \mathbb{R}^{h \times w}$, where each pixel $(x, y)$ quantifies the significance of corresponding patches in the original image, we proceed to establish a precise mapping to our intended feature map (Fig.~\ref{fig:2}). This mapping ensures the preservation of spatial relationships within the feature map.

Consider a target feature map $\mathcal{F} \in \mathbb{R}^{h' \times w' \times K}$, with each pixel $(x', y')$ encapsulating a feature vector from a specific image patch. To align $\mathcal{R}$ with $\mathcal{F}$, we employ an upsampling operation $U: \mathbb{R}^{h \times w} \rightarrow \mathbb{R}^{h' \times w'}$, transforming $\mathcal{R}$ into the Patch Significance Map (PSM) $\mathcal{P} \in \mathbb{R}^{h' \times w'}$. Consequently, each element of $\mathcal{P}$ directly corresponds to the patch significance within $\mathcal{F}$ for the given discriminative task.
The next step involves utilizing $\mathcal{P}$ to discern and prune non-essential patch features from $\mathcal{F}$. By sorting the values within $\mathcal{P}$, we acquire a pruning mask $\mathcal{M}$:
\begin{equation}
    \mathcal{M} = \mbox{reshape}(\mbox{argsort} (\mbox{flatten}(\mathcal{P}))), \mathcal{M} \in \mathbb{R}^{h' \times w'},
\end{equation}
enabling batch-wise patch pruning. Specifically, we identify the indices corresponding to the top-\textit{$z\%$} patches as per $\mathcal{M}$, facilitating the retention of only the most salient patches in $\mathcal{F}$, thereby enhancing computational efficiency in subsequent processing stages.

As part of our discussion on various architectures, we detail the application of this patch reduction technique. For enhanced visualization of the PSM, we apply \textit{min-max} normalization to $\mathcal{P}$, adjusting for outliers and scaling the significance scores to visually depict the importance of different patches.

\vspace{-2mm}
\subsection{Integrating PSM with Vision Transformers}
\begin{figure}[t]
\centering
\includegraphics[width= 1.0\textwidth]{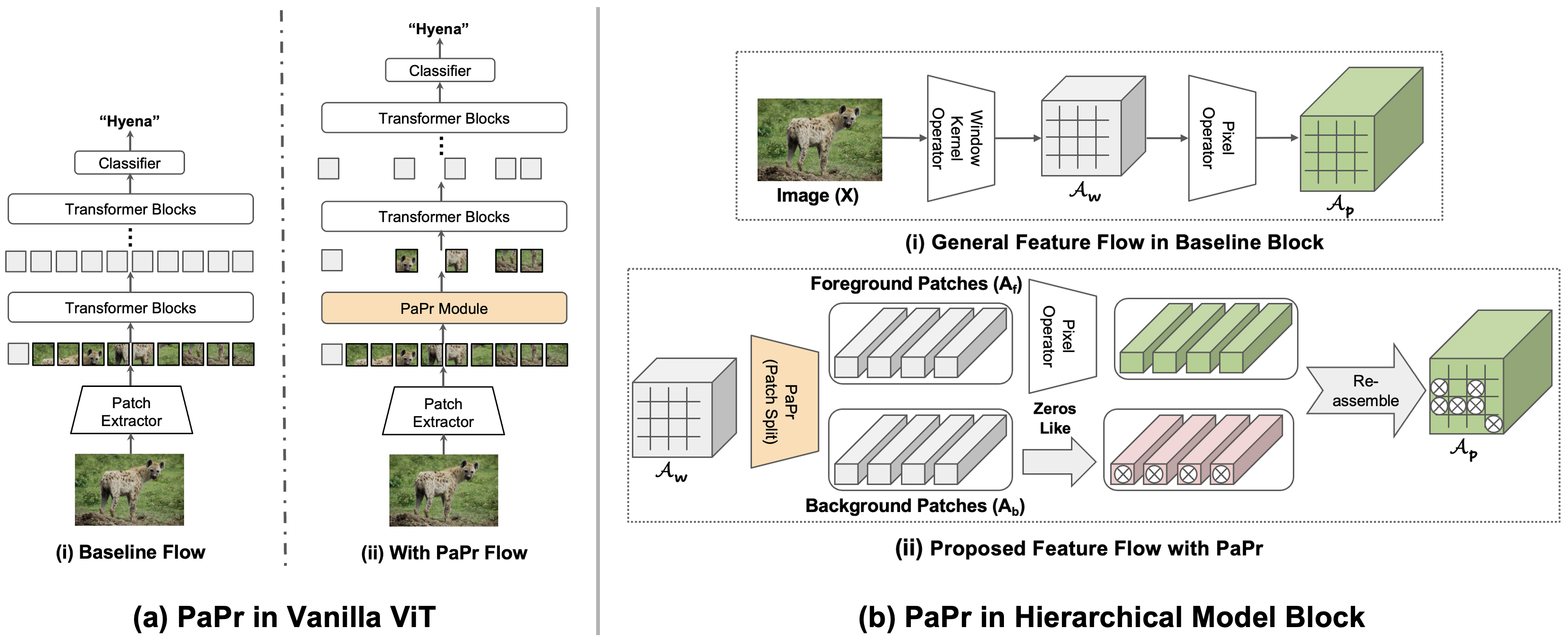}
\caption{(a) In vanilla ViT, PaPr operates right after the patch extractor module. Hence, all transformer blocks can operate only with the most discriminative patches. (b) Hierarchical model blocks comprise of window based kernel operator (\textit{e.g.} Conv$k\times k$/local attention), followed by pixel operator (\textit{e.g.}, linear layer, Conv1x1). Pixel operator consumes more than 60\% of total computation. PaPr is used to split the foreground patches to be operated with pixel operator. Background patches are zero-ed out, and finally, re-assembled with foreground output patches.}
\label{fig:3}
\vspace{-4mm}
\end{figure}

The ViT processes an input image $X \in \mathbb{R}^{H \times W \times 3}$ by extracting $N$-dimensional patch token features $X_p \in \mathbb{R}^{N \times d}$ from a $(k \times k)$ patch window using strided convolutions, where $N = HW/{k^2}$ (Fig.~\ref{fig:3}). Positional embeddings are then added to these patch tokens $X_p$ to retain spatial information, followed by the application of successive cross-attention mechanisms. For classification tasks, the model either introduces a class token $t_{cls} \in \mathbb{R}^{d}$ or utilizes the mean of the output patch tokens.

The encoder employs multi-headed cross-attention (MHA) on the patch token embeddings $X_p$, with each MHA block consisting of multiple linear layers and a cross-attention layer. The computational complexity of these operations suggests that reducing the number of tokens $N$ can significantly decrease computational demands. A key strength of the ViT architecture is its adaptability to varying numbers of tokens $N$, though at a higher computational cost.
Prior work~\cite{dynamicvit, Avit, tome} has primarily aimed at reducing patch tokens within intermediate encoder layers, often requiring additional training due to the variability in token representations across different architectures. These methods typically achieve only gradual token reduction, limited by the distributed nature of information across tokens.

In contrast, our approach seeks to eliminate redundant tokens early in the processing pipeline, immediately following initial patch extraction. This strategy offers multiple advantages: (1) Leveraging the convolutional nature of the initial patch extraction allows for a direct mapping between our PSM $\mathcal{P}$ and the patch tokens $X_p$, obviating the need for additional mechanisms to determine token redundancy. (2) It facilitates generalizability across transformer models without the need for retraining, potentially accelerating training by concentrating on essential patches. (3) It permits seamless integration of existing intermediate layer token reduction techniques following our initial patch pruning process.

\vspace{-2mm}
\subsection{Integrating PSM with Hierarchical Models}

ConvNets and hybrid transformers, such as Swin~\cite{swin}, primarily operate with window-based local operations in contrast to full-attention based operation in vanilla transformers. Such window-based local processing makes patch pruning particularly complicated, in contrast to vanilla transformers. However, these models maintain the location property of representative image patches all through the network, which leaves the door open to prune redundant patches. Nevertheless, the window-based operation is particularly difficult to prune.

In general, each block of hierarchical layers consists of window-based spatial kernel/attention operators, followed by pixel operators (usually, modeled with $1\times1$ convolutional layers or linear layers).
In contrast to windowed convolutions or cross-attention, these pixel operators are particularly suitable for patch pruning. 
Interestingly, in the SOTA ConvNets and hybrid transformers, more than 60\% of total computations are performed with such pixel operators ($63.3\%$ in Swin~\cite{swin}, $96.2\%$ in ConvNext~\cite{convnet2020}). 
Based on the patch significance map from PaPr, we only use the pixel operator on most significant patches, and simply perform zero-padding on the remaining patch regions.
The zero padding operations are performed to mostly recover the feature map shape to be used with subsequent spatial operators. Hence, the speed-up is mostly achieved by eliminating redundant computations in pixel operators (See Fig.~\ref{fig:3}). 

Let's assume that the feature map before applying pixel operators is given by $\mathcal{A}_w$. Based on the PSM $\mathcal{P}$, we initially identify the foreground and background pixel features as $A_f$, and $A_b$, respectively. Later, the pixel operator layers are applied on foreground pixels $A_f$, and a zero-padded representation is used for the background pixels $A_b$. Finally, the output representation $\mathcal{A}_c$ is reassembled with modulated foreground pixels $A_f$ and zero-padded background pixels $A_z$, given by Eq.~(\ref{eq:hierarchical}):
\begin{equation}
    \begin{aligned}
    A_f, A_{b} = \mbox{Split}(\mathcal{A}_w, \mathcal{P}), \\ 
    A_z = \mbox{Zeros}(A_{b}),    A_{f} = \mbox{Linear}(A_f), \\
    \mathcal{A}_c= \mbox{Reassemble}(A_f, A_z).
    \label{eq:hierarchical}
    \end{aligned}
\end{equation}

\vspace{-4mm}
\section{Image Experiments}
\vspace{-2mm}

\begin{figure}[t]
\centering
\begin{minipage}[b]{0.48\linewidth}
\centering
\includegraphics[width= \linewidth]{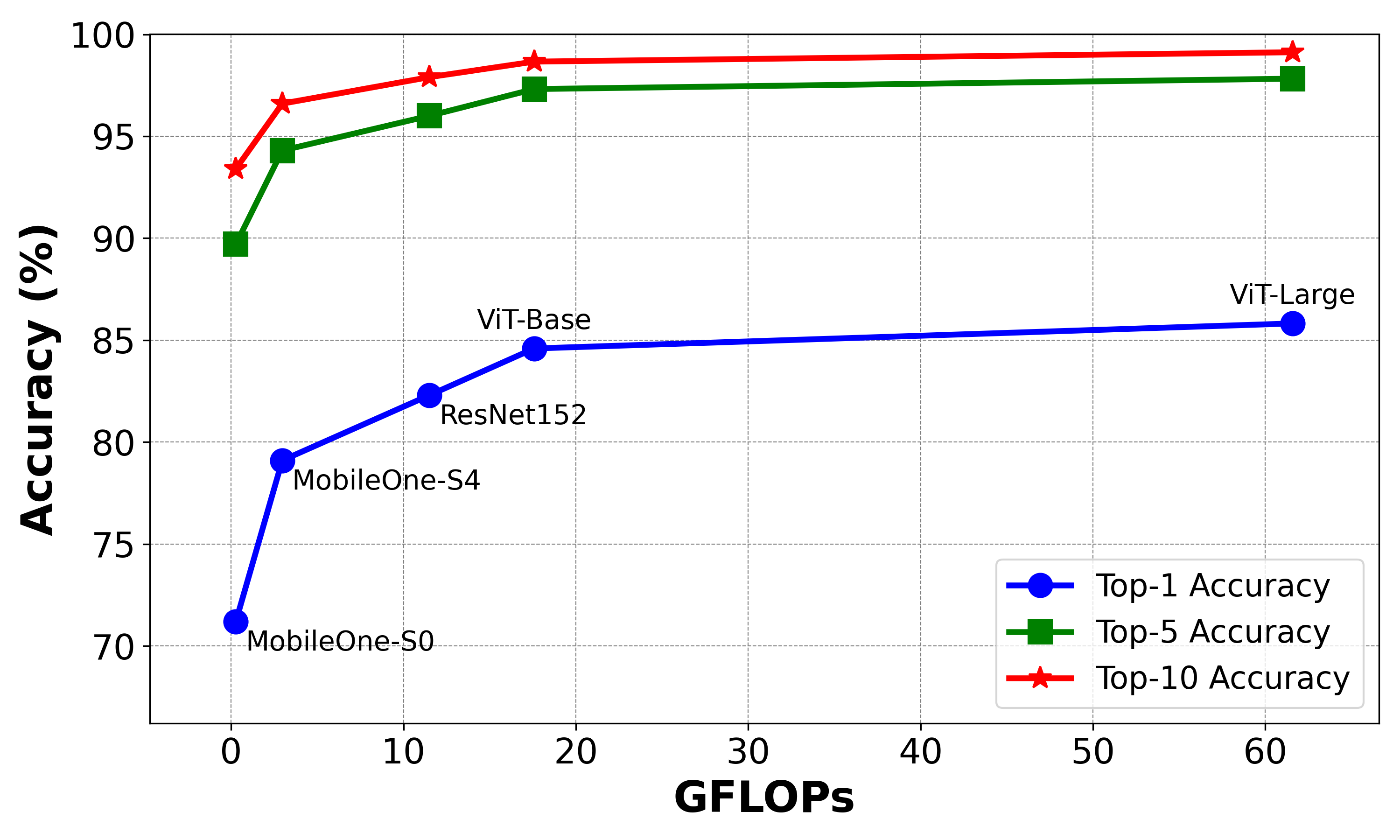}
\caption{ImageNet-1k evaluation for varying top-k accuracy targets. The accuracy gain with bigger model largely shrinks, as $k$ increases. This suggests shallower ConvNets have understanding of object locations and visual property, despite their  lower top-1 accuracy.}
\label{fig:4}
\end{minipage}
\hspace{1pt}
\begin{minipage}[b]{0.48\linewidth}
\centering
\includegraphics[width= \linewidth]{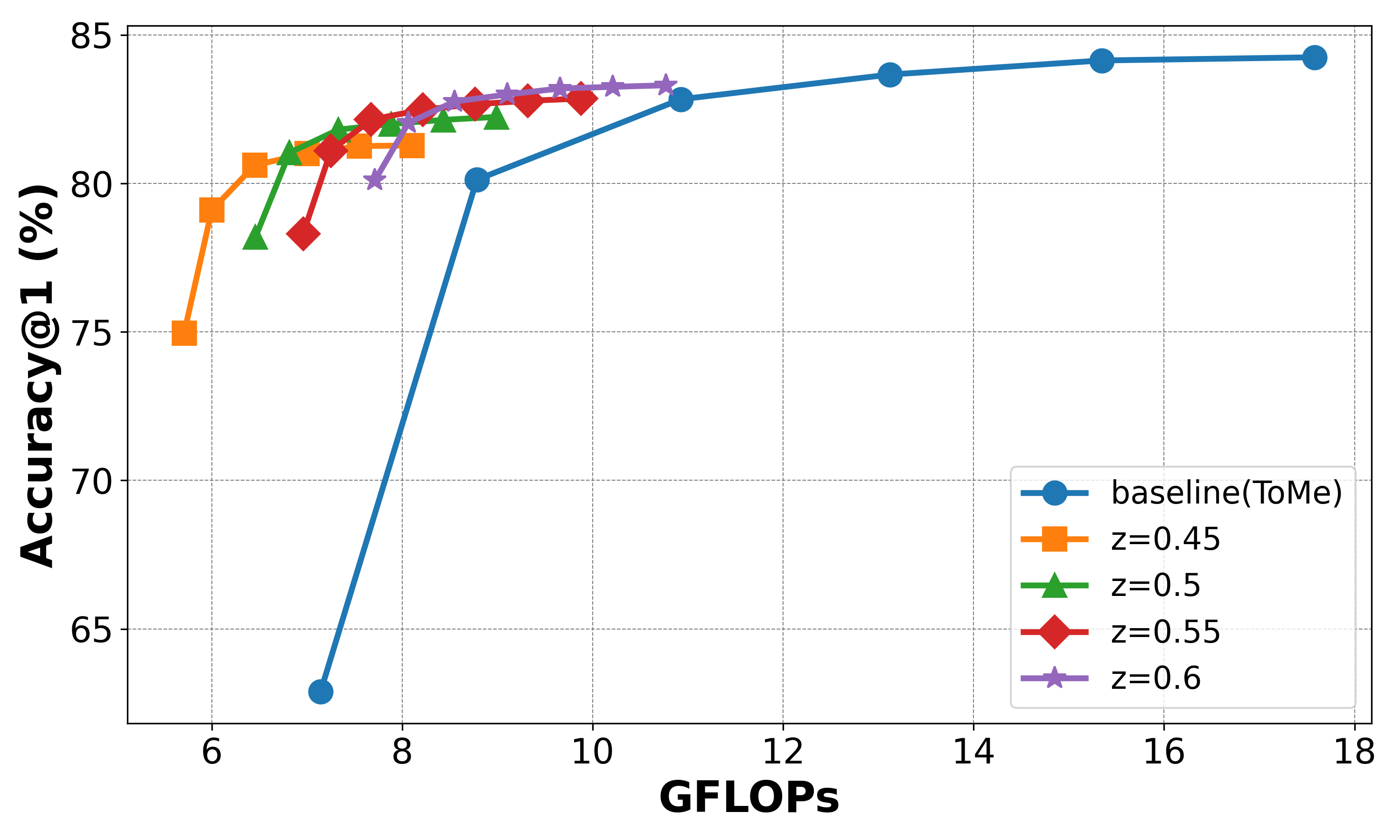}
\caption{Integrating PaPr with ToMe~\cite{tome}. We use the Augreg pretrained ViT-B-16 architecture as the baseline. We sweep token merging ratio (r) for different pruning ratio (z). Integration of PaPr achieves Pareto-optimal performance, thus, PaPr can enhance existing patch reduction methods.}
\label{fig:5}
\end{minipage}
\vspace{-3mm}
\end{figure}

\begin{table}[t]
\centering

\begin{minipage}{0.5\linewidth}\centering
\caption{Performance comparison on AugReg models. PaPr achieves the best performance, while operating with token merging. $z$ denotes patch keeping ratio in PaPr.}
\label{tab1}
\vspace{-1mm}
\scalebox{0.7}{
\begin{tabular}{ccccc}
\toprule
\textbf{Models}            & \textbf{Methods}                              & \textbf{Acc1}                      & \textbf{GFLOPs}                      & \textbf{Img/s}                 \\
\midrule
                           & Baseline                                      & 81.39                                      & 4.61                                      & 975                                        \\
                           & ToMe~\cite{tome}                                          & 76.96                                  & 2.29                                   & 1978                                  \\
                           & TokenFusion~\cite{tokenfusion}                                   & 77.12                                      & 2.29                                      & 1982                                        \\
                           & GTP-ViT~\cite{xu2024gtp}                                       & 71.03                                      & 2.31                                      & 1970                                        \\
                           & \cellcolor[HTML]{EFEFEF}PaPr (z=0.45)         & \cellcolor[HTML]{EFEFEF}76.21          & \cellcolor[HTML]{EFEFEF}2.29           & \cellcolor[HTML]{EFEFEF}1988          \\
                           & \cellcolor[HTML]{EFEFEF}PaPr (z=0.55) (+ToMe) & \cellcolor[HTML]{EFEFEF}\textbf{77.76} & \cellcolor[HTML]{EFEFEF}2.19           & \cellcolor[HTML]{EFEFEF}2073          \\
\multirow{-7}{*}{ViT-S-16} & \cellcolor[HTML]{EFEFEF}PaPr (z=0.5) (+ToMe)  & \cellcolor[HTML]{EFEFEF}76.27          & \cellcolor[HTML]{EFEFEF}\textbf{1.97}  & \cellcolor[HTML]{EFEFEF}\textbf{2315} \\
\midrule
                           & Baseline                                      & 84.59                                      & 17.59                                      & 307                                        \\
                           & ToMe~\cite{tome}                                          & 80.38                                  & 8.78                                   & 615                                   \\
                           & TokenFusion~\cite{tokenfusion}                                   & 80.7                                      & 8.78                                      & 618                                        \\
                           & GTP-ViT~\cite{xu2024gtp}                                       & 80.98                                      & 8.78                                      & 610                                        \\
                           & \cellcolor[HTML]{EFEFEF}PaPr (z=0.5)           & \cellcolor[HTML]{EFEFEF}82.11         & \cellcolor[HTML]{EFEFEF}8.98           & \cellcolor[HTML]{EFEFEF}605           \\
                           & \cellcolor[HTML]{EFEFEF}PaPr (z=0.55) (+ToMe) & \cellcolor[HTML]{EFEFEF}\textbf{82.34} & \cellcolor[HTML]{EFEFEF}8.21           & \cellcolor[HTML]{EFEFEF}660           \\
\multirow{-7}{*}{ViT-B-16} & \cellcolor[HTML]{EFEFEF}PaPr (z=0.5) (+ToMe)  & \cellcolor[HTML]{EFEFEF}80.88          & \cellcolor[HTML]{EFEFEF}\textbf{6.82}  & \cellcolor[HTML]{EFEFEF}\textbf{785}  \\
\midrule
                           & Baseline                                      & 85.82                                      & 61.61                                      & 91                                        \\
                           & ToMe~\cite{tome}                                          & 83.5                                   & 30.99                                      & 180                                        \\
                           & TokenFusion~\cite{tokenfusion}                                   & 83.91                                      & 30.99                                      & 182                                        \\
                           & GTP-ViT~\cite{xu2024gtp}                                       & 81.56                                      & 30.99                                      & 181                                        \\
                           & \cellcolor[HTML]{EFEFEF}PaPr (z=0.5)          & \cellcolor[HTML]{EFEFEF}83.87          & \cellcolor[HTML]{EFEFEF}30.83          & \cellcolor[HTML]{EFEFEF}183           \\
                           & \cellcolor[HTML]{EFEFEF}PaPr (z=0.55) (+ToMe) & \cellcolor[HTML]{EFEFEF}\textbf{83.99} & \cellcolor[HTML]{EFEFEF}26.33          & \cellcolor[HTML]{EFEFEF}210           \\
\multirow{-7}{*}{ViT-L-16} & \cellcolor[HTML]{EFEFEF}PaPr (z=0.55) (+ToMe) & \cellcolor[HTML]{EFEFEF}83.5           & \cellcolor[HTML]{EFEFEF}\textbf{25.12} & \cellcolor[HTML]{EFEFEF}\textbf{224}    \\
\bottomrule
\end{tabular}}
\end{minipage}
\hspace{1pt}
\begin{minipage}{0.46\linewidth}\centering
\caption{Performance comparison on MAE models. Since MAE uses masked pretraining, PaPr is particularly suitable for MAE inference. PaPr achieves significantly higher performance than others. $z$ denotes patch keeping ratio in PaPr.}
\label{tab2}
\vspace{-2mm}
\scalebox{0.72}{
\begin{tabular}{ccccc}
\toprule
\textbf{Models}            & \textbf{Methods}                     & \textbf{Acc1}                      & \textbf{GFLOPs}                        & \textbf{Img/s}                          \\
\midrule
                           & Baseline                             & 83.74                                      & 17.59                                      & 307                                       \\
                           & ToMe~\cite{tome}                                 & 78.82                                  & 8.78                                   & 615                                  \\
                           & TokenFusion~\cite{tokenfusion}                          & 79.23                                      & 8.78                                      & 618                                       \\
                           & GTP-ViT~\cite{xu2024gtp}                              & 79.14                                      & 8.78                                      & 610                                       \\
                           & \cellcolor[HTML]{EFEFEF}PaPr (z=0.5) & \cellcolor[HTML]{EFEFEF}\textbf{82.4}  & \cellcolor[HTML]{EFEFEF}8.98           & \cellcolor[HTML]{EFEFEF}605          \\
\multirow{-6}{*}{ViT-B-16} & \cellcolor[HTML]{EFEFEF}PaPr (z=0.4) & \cellcolor[HTML]{EFEFEF}81.4           & \cellcolor[HTML]{EFEFEF}\textbf{7.72}  & \cellcolor[HTML]{EFEFEF}\textbf{700} \\
\midrule
                           & Baseline                             & 85.95                                      & 61.61                                      & 91                                       \\
                           & ToMe~\cite{tome}                                 & 84.24                                  & 30.99                                  & 180                                  \\
                           & TokenFusion~\cite{tokenfusion}                          & 84.33                                      & 30.99                                      & 182                                       \\
                           & GTP-ViT~\cite{xu2024gtp}                              & 84.15                                      & 30.99                                      & 181                                       \\
                           & \cellcolor[HTML]{EFEFEF}PaPr (z=0.5) & \cellcolor[HTML]{EFEFEF}\textbf{85.06} & \cellcolor[HTML]{EFEFEF}30.83          & \cellcolor[HTML]{EFEFEF}183          \\
\multirow{-6}{*}{ViT-L-16} & \cellcolor[HTML]{EFEFEF}PaPr (z=0.4) & \cellcolor[HTML]{EFEFEF}84.76          & \cellcolor[HTML]{EFEFEF}\textbf{27.72} & \cellcolor[HTML]{EFEFEF}\textbf{201} \\
\midrule
                           & Baseline                             & 86.89                                      & 167.4                                      & 36                                       \\
                           & ToMe~\cite{tome}                                 & 85.48                                  & 82.53                                  & 72                                   \\
                           & TokenFusion~\cite{tokenfusion}                          & 85.71                                      & 82.53                                      & 73                                       \\
                           & GTP-ViT~\cite{xu2024gtp}                              & 85.54                                      & 82.53                                      & 71                                       \\
                           & \cellcolor[HTML]{EFEFEF}PaPr (z=0.5) & \cellcolor[HTML]{EFEFEF}\textbf{86.4}  & \cellcolor[HTML]{EFEFEF}83.04          & \cellcolor[HTML]{EFEFEF}71           \\
\multirow{-6}{*}{ViT-H-16} & \cellcolor[HTML]{EFEFEF}PaPr (z=0.4) & \cellcolor[HTML]{EFEFEF}86.13          & \cellcolor[HTML]{EFEFEF}\textbf{74.59} & \cellcolor[HTML]{EFEFEF}\textbf{81}                \\
\bottomrule
\end{tabular}}
\end{minipage}
\
\end{table}

\begin{table}[t]
\centering
\begin{minipage}[b]{0.48\linewidth}\centering
\caption{Performance analysis on class-token free ViT models. PaPr performance gain is not limited to specific architectures.}
\label{tab4}
\scalebox{0.74}{
\begin{tabular}{ccccc}
\toprule
\textbf{Model}                                                                      & \textbf{Method}              & \textbf{Acc1}         & \textbf{GFLOPs}            & \textbf{Img/s}            \\
\midrule
                                                                                    & Baseline                     & 84.33                         & 10.58                         & 484                         \\
                                                                                    & GTP-ViT~\cite{xu2024gtp}                      & 80.5                         & 5.73                         & 880                         \\
                                                                                    & ToMe~\cite{tome}                         & 80.21                         & 5.73                         & 890                         \\
\multirow{-4}{*}{\begin{tabular}[c]{@{}c@{}}ViT-Medium\\ GAP-16\\ 256\end{tabular}} & \cellcolor[HTML]{EFEFEF}PaPr (z=0.5) & \cellcolor[HTML]{EFEFEF}\textbf{81.8} & \cellcolor[HTML]{EFEFEF} \textbf{5.5} & \cellcolor[HTML]{EFEFEF}\textbf{932} \\
\midrule
                                                                                    & Baseline                     & 85.6                         & 26.06                         & 182                         \\
                                                                                    & GTP-ViT~\cite{xu2024gtp}                      & 81.95                         & 13.59                         &   340                       \\
                                                                                    & ToMe~\cite{tome}                         & 82.5                         & 13.59                         & 344                         \\
\multirow{-4}{*}{\begin{tabular}[c]{@{}c@{}}ViT-Medium\\ GAP-16\\ 384\end{tabular}}   & \cellcolor[HTML]{EFEFEF}PaPr & \cellcolor[HTML]{EFEFEF}\textbf{83.95} & \cellcolor[HTML]{EFEFEF}\textbf{12.94} & \cellcolor[HTML]{EFEFEF}\textbf{360}  \\
\bottomrule
\end{tabular}}
\end{minipage}
\hspace{1pt}
\begin{minipage}[b]{0.48\linewidth}\centering
\caption{Training performance analysis on DeIT-s. PaPr achieves competitive performance with higher training speed.}
\label{tab3}
\scalebox{0.7}{
\begin{tabular}{cccccc}
\toprule
                                   &                                     &                                  &                                  &                                                                                   &                                                                                   \\
\multirow{-2}{*}{\textbf{Methods}} & \multirow{-2}{*}{\textbf{Acc1}} & \multirow{-2}{*}{\textbf{GFLOPs}} & \multirow{-2}{*}{\textbf{Im/s}} & \multirow{-2}{*}{\textbf{\begin{tabular}[c]{@{}c@{}}Batch \\ Proc.\end{tabular}}} & \multirow{-2}{*}{\textbf{\begin{tabular}[c]{@{}c@{}}Train \\ Speed\end{tabular}}} \\
\midrule
{\color[HTML]{9B9B9B} Baseline}    & {\color[HTML]{9B9B9B} 79.8}         & {\color[HTML]{9B9B9B} 4.6}       & {\color[HTML]{9B9B9B} 960}       & {\color[HTML]{9B9B9B} \cmark}                                                          & {\color[HTML]{9B9B9B} 1x}                                                                                \\
DynamicViT~\cite{dynamicvit}                         & 79.3                                & 2.9                              & 1510                             & \xmark                                                                                 & 1x                                                                                \\
A-ViT~\cite{Avit}                              & 78.6                                & 2.9                              & -                                & \xmark                                                                                 & 1x                                                                                \\
ATS~\cite{ats}                                & \textbf{79.5}                                   & 2.9                              & 1512                                & -                                                                                 & 1x                                                                                \\
SP-ViT~\cite{spvit}                             & 79.3                                & 2.6                              & -                                & \xmark                                                                                 & 1x                                                                                \\
ToMe~\cite{tome}                               & 79.4                       & 2.7                              & 1575                             & \cmark                                                                                 & 1.5x                                                                              \\
TokenFusion~\cite{tokenfusion}                            & \textbf{79.5}                                & 2.7                              & 1580                                & \cmark                                                                                 & {1.5x}                                                                              \\
\rowcolor[HTML]{EFEFEF} 
PaPr (z=0.55)                      & 79.2                                & 2.7                              & \textbf{1585}                    & \cmark                                                                                 & \textbf{1.6x}                                                                             \\
\bottomrule
\end{tabular}}
\end{minipage}
\vspace{-4mm}
\end{table}

\begin{table}[t]
\centering

\begin{minipage}[b]{0.48\linewidth}
\centering
\caption{Performance comparison on ConvNext CNN models. PaPr can achieve competitive performance without training, and can seamlessly adapt to model updates.}
\label{tab5}
\scalebox{0.63}{
\begin{tabular}{cccccc}
\toprule
\textbf{Models}                                                                & \textbf{Methods}                      & \textbf{Train-Free}          & \textbf{Acc1}             & \textbf{GFLOPs}                & \textbf{Img/s}                          \\
\midrule
                                                                               & {\color[HTML]{9B9B9B} Baseline}       & {\color[HTML]{9B9B9B} N/A} & {\color[HTML]{9B9B9B} 83.84}  & {\color[HTML]{9B9B9B} 15.38}  & {\color[HTML]{9B9B9B} 265}           \\
                                                                               & DynCNN~\cite{dyncnn}                            & \xmark                          & {83.08}                & 10.21                         & 375                                  \\
\multirow{-3}{*}{\begin{tabular}[c]{@{}c@{}}ConvNeXt\\ Base-1k\end{tabular}}   & \cellcolor[HTML]{EFEFEF}PaPr (z=0.65) & \cellcolor[HTML]{EFEFEF}\cmark  & \cellcolor[HTML]{EFEFEF}82.75 & \cellcolor[HTML]{EFEFEF}10.42 & \cellcolor[HTML]{EFEFEF}\textbf{395} \\
\midrule
                                                                               & {\color[HTML]{9B9B9B} Baseline}       & {\color[HTML]{9B9B9B} N/A} & {\color[HTML]{9B9B9B} 85.81}  & {\color[HTML]{9B9B9B} 15.38}  & {\color[HTML]{9B9B9B} 265}           \\
\multirow{-2}{*}{\begin{tabular}[c]{@{}c@{}}ConvNeXt\\ Base-22k\end{tabular}}  & \cellcolor[HTML]{EFEFEF}PaPr (z=0.65) & \cellcolor[HTML]{EFEFEF}\cmark  & \cellcolor[HTML]{EFEFEF}84.27 & \cellcolor[HTML]{EFEFEF}10.42 & \cellcolor[HTML]{EFEFEF}395          \\
\midrule
                                                                               & {\color[HTML]{9B9B9B} Baseline}       & {\color[HTML]{9B9B9B} N/A} & {\color[HTML]{9B9B9B} 84.31}  & {\color[HTML]{9B9B9B} 34.4}   & {\color[HTML]{9B9B9B} 135}            \\
\multirow{-2}{*}{\begin{tabular}[c]{@{}c@{}}ConvNeXt\\ Large-1k\end{tabular}}  & \cellcolor[HTML]{EFEFEF}PaPr (z=0.65) & \cellcolor[HTML]{EFEFEF}\cmark  & \cellcolor[HTML]{EFEFEF}83.26 & \cellcolor[HTML]{EFEFEF}22.9  & \cellcolor[HTML]{EFEFEF}203           \\
\midrule
                                                                               & {\color[HTML]{9B9B9B} Baseline}       & {\color[HTML]{9B9B9B} N/A} & {\color[HTML]{9B9B9B} 86.61}  & {\color[HTML]{9B9B9B} 34.4}   & {\color[HTML]{9B9B9B} 135}            \\
\multirow{-2}{*}{\begin{tabular}[c]{@{}c@{}}ConvNeXt\\ Large-22k\end{tabular}} & \cellcolor[HTML]{EFEFEF}PaPr (z=0.65) & \cellcolor[HTML]{EFEFEF}\cmark  & \cellcolor[HTML]{EFEFEF}\textbf{85.67} & \cellcolor[HTML]{EFEFEF}22.9  & \cellcolor[HTML]{EFEFEF}203                     \\
\bottomrule
\end{tabular}}

\end{minipage}
\hspace{1pt}
\begin{minipage}[b]{0.48\linewidth}\centering
\caption{Performance comparison on Swin hybrid transformer models. PaPr can adapt to much bigger models with higher operating resolutions without training.}
\label{tab6}
\scalebox{0.63}{
\begin{tabular}{cccccc}
\toprule
\textbf{Models}                                                                  & \textbf{Methods}                      & \textbf{Train-Free}          & \textbf{Acc1}             & \textbf{GFLOPs}                & \textbf{Img/s}                 \\
\midrule
                                                                                 & {\color[HTML]{9B9B9B} Baseline}       & {\color[HTML]{9B9B9B} N/A} & {\color[HTML]{9B9B9B} 83.42}  & {\color[HTML]{9B9B9B} 15.47}  & {\color[HTML]{9B9B9B} 258}  \\
                                                                                 & DynSwin~\cite{dyncnn}                            & \xmark                          & {83.18} 
                                                                                 & 12.1                          & \textbf{327}                \\
\multirow{-3}{*}{\begin{tabular}[c]{@{}c@{}}Swin\\ B-1k\end{tabular}}         & \cellcolor[HTML]{EFEFEF}PaPr (z=0.65) & \cellcolor[HTML]{EFEFEF}\cmark  & \cellcolor[HTML]{EFEFEF}81.7  & \cellcolor[HTML]{EFEFEF}12.25 & \cellcolor[HTML]{EFEFEF}325 \\
\midrule
                                                                                 & {\color[HTML]{9B9B9B} Baseline}       & {\color[HTML]{9B9B9B} N/A} & {\color[HTML]{9B9B9B} 85.16}  & {\color[HTML]{9B9B9B} 15.47}  & {\color[HTML]{9B9B9B} 258}  \\
\multirow{-2}{*}{\begin{tabular}[c]{@{}c@{}}Swin\\ B-22k\end{tabular}}        & \cellcolor[HTML]{EFEFEF}PaPr (z=0.65) & \cellcolor[HTML]{EFEFEF}\cmark  & \cellcolor[HTML]{EFEFEF}82.27 & \cellcolor[HTML]{EFEFEF}12.25 & \cellcolor[HTML]{EFEFEF}325 \\
\midrule
                                                                                 & {\color[HTML]{9B9B9B} Baseline}       & {\color[HTML]{9B9B9B} N/A} & {\color[HTML]{9B9B9B} 86.25}  & {\color[HTML]{9B9B9B} 34.53}  & {\color[HTML]{9B9B9B} 135}  \\
\multirow{-2}{*}{\begin{tabular}[c]{@{}c@{}}Swin\\ L-22k\end{tabular}}       & \cellcolor[HTML]{EFEFEF}PaPr (z=0.65) & \cellcolor[HTML]{EFEFEF}\cmark  & \cellcolor[HTML]{EFEFEF}84.53 & \cellcolor[HTML]{EFEFEF}26.96 & \cellcolor[HTML]{EFEFEF}175 \\
\midrule
                                                                                 & {\color[HTML]{9B9B9B} Baseline}       & {\color[HTML]{9B9B9B} N/A} & {\color[HTML]{9B9B9B} 87.25}  & {\color[HTML]{9B9B9B} 104.08} & {\color[HTML]{9B9B9B} 42}    \\
\multirow{-2}{*}{\begin{tabular}[c]{@{}c@{}}Swin\\ L-22k-384\end{tabular}} & \cellcolor[HTML]{EFEFEF}PaPr (z=0.65) & \cellcolor[HTML]{EFEFEF}\cmark  & \cellcolor[HTML]{EFEFEF}\textbf{86.47} & \cellcolor[HTML]{EFEFEF}81.44 & \cellcolor[HTML]{EFEFEF}54  \\
\bottomrule
\end{tabular}}
\end{minipage}
\end{table}

\begin{figure}[t]
\centering

\begin{minipage}{0.45\linewidth}{
    \centering
    \captionof{table}{Sweeping  ConvNet proposal model in PaPr with $z=0.5$. PaPr can achieve similar performance irrespective of proposal model size. Thus, PaPr can use a much smaller and faster proposal model to speed-up larger models.}
    \label{tab7}
    \scalebox{0.72}{
    \begin{tabular}{ccccc}
        \toprule
        \multicolumn{2}{c}{\textbf{Proposal}} & \multicolumn{2}{c}{\textbf{Accuracy1(\%)}} \\
        \cmidrule(lr){1-2} \cmidrule(lr){3-4}
        \textbf{Model}    & \textbf{GFLOPs}   & \textbf{ViT-B-16} & \textbf{ViT-L-16} \\
        \midrule
        ResNet-18~\cite{resnet}         & 1.81              & 81.1              & 83.84             \\
        ResNet-50~\cite{resnet}          & 4.09              & 82.33             & 84.09             \\
        ResNet-152~\cite{resnet}         & 11.51             & \textbf{82.51}             & 84.08             \\
        \midrule
        MobileOne-S0~\cite{mobileone}      & 0.27              & 82.24             & 84.06             \\
        MobileOne-S2~\cite{mobileone}      & 1.35              & 82.28             & 84.16             \\
        MobileOne-S4~\cite{mobileone}      & 2.98              & 82.35             & \textbf{84.32}              \\
        \bottomrule
    \end{tabular}}
}\end{minipage}
\hspace{1em}
\begin{minipage}{0.45\linewidth}{
\includegraphics[width= \linewidth]{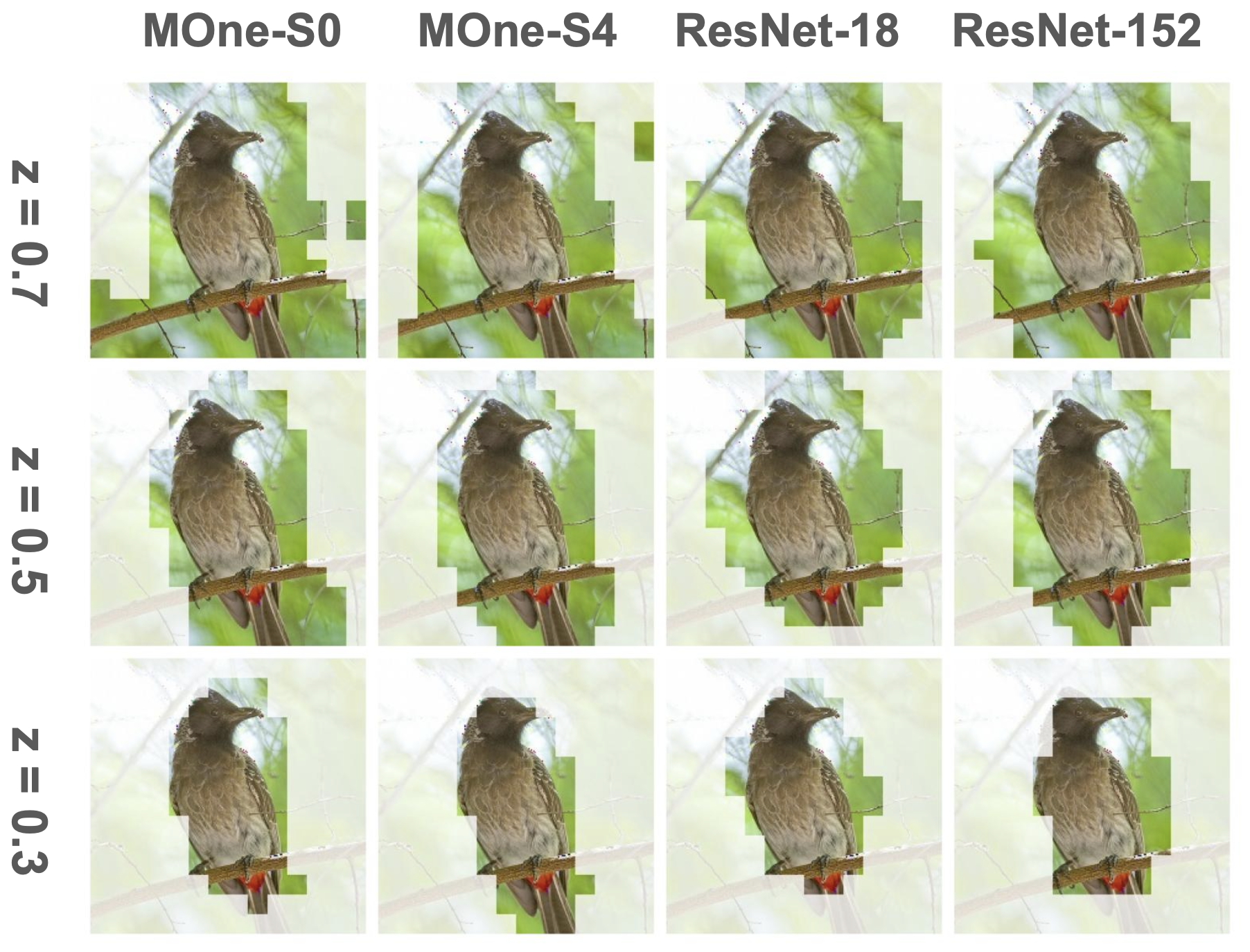}
    \centering
    \captionof{figure}{Proposal models locate similar patches with PaPr for different pruning ratio (\textit{z}) irrespective of model size.}
    \label{fig:6}
}\end{minipage}
\vspace{-2mm}
\end{figure}

\begin{figure}[t]
\centering
\includegraphics[width= 0.95\textwidth]{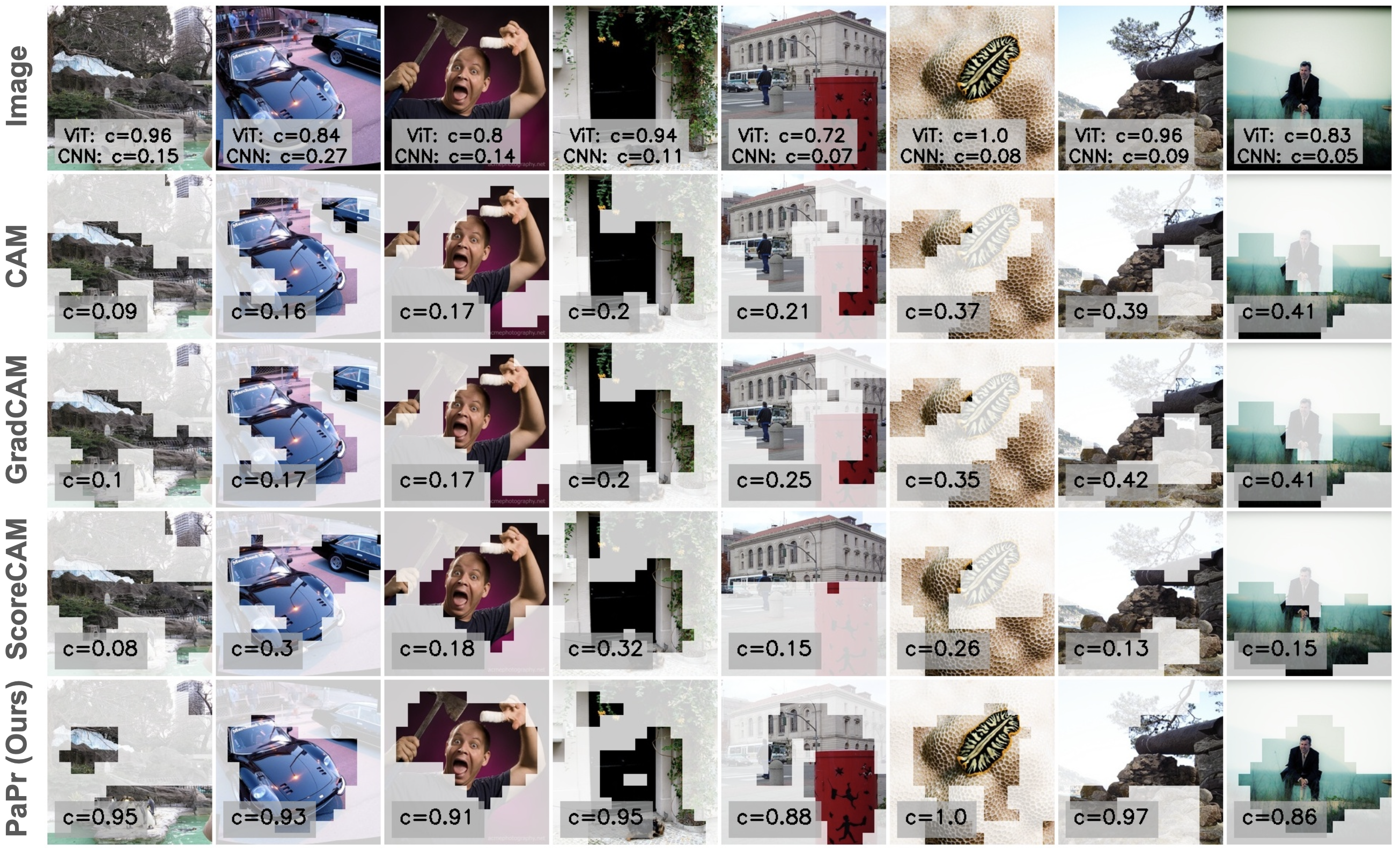}
\caption{Robustness of PaPr compared to CAM based methods. PaPr can perform even when the ConvNet proposal confidence (c) is very low. In contrast, existing CAM based methods fail in such cases, despite being significantly slower while not enabling batch processing and use of gradients in some cases. Moreover, PaPr can even enhance the ViT confidence in several challenging scenarios by removing redundant patches.}
\label{fig:7}
\vspace{-5mm}
\end{figure}

\begin{table}[t]
\centering
\begin{minipage}[b]{0.5\linewidth}\centering
\caption{
Training-free performance comparison on Kinetics-400 video evaluation.
Each video has $16\times224^2$ input size. PaPr achieves significantly better performance for reducing spatio-temporal redundancy in videos.}
\label{tab8}
\scalebox{0.72}{
\begin{tabular}{ccccc}
\toprule
\textbf{Model}                                                        & \textbf{Method}                      & \textbf{Acc1}                 & \textbf{GFLOPs}            & \textbf{Views}              \\
\midrule
XViT                                                                  & ATS~\cite{ats}                                  & 80.0                          & 259                        & 1x3                         \\
\midrule
TimeSformer-L                                                         & ATS~\cite{ats}                                  & 80.5                          & 3510                       & 1x3                         \\
\midrule
                                                                      & {\color[HTML]{9B9B9B} Baseline}      & {\color[HTML]{9B9B9B} 81.21}  & {\color[HTML]{9B9B9B} 180} & {\color[HTML]{9B9B9B} 3x5}  \\
                                                                      & \cellcolor[HTML]{EFEFEF}PaPr(z=0.45) & \cellcolor[HTML]{EFEFEF}\textbf{81.18} & \cellcolor[HTML]{EFEFEF}76  & \cellcolor[HTML]{EFEFEF}3x5                         \\
\multirow{-3}{*}{\begin{tabular}[c]{@{}c@{}}ViT-B\\ MAE\end{tabular}} & \cellcolor[HTML]{EFEFEF}PaPr(z=0.35) & \cellcolor[HTML]{EFEFEF}80.15 & \cellcolor[HTML]{EFEFEF}\textbf{59}  & \cellcolor[HTML]{EFEFEF}3x5                         \\
\midrule
                                                                      & {\color[HTML]{9B9B9B} Baseline}      & {\color[HTML]{9B9B9B} 85.26}  & {\color[HTML]{9B9B9B} 598} & {\color[HTML]{9B9B9B} 3x5}  \\
                                                                      & ToMe~\cite{tome}                                 & 84.5                          & 281                        & 1x10                        \\
                                                                      & \cellcolor[HTML]{EFEFEF}PaPr(z=0.5)  & \cellcolor[HTML]{EFEFEF}\textbf{85.12}     & \cellcolor[HTML]{EFEFEF}275  & \cellcolor[HTML]{EFEFEF}3x5 \\
                                                                      & ToMe~\cite{tome}                                 & 82.5                          & 184                        & 1x10                        \\
\multirow{-5}{*}{\begin{tabular}[c]{@{}c@{}}ViT-L\\ MAE\end{tabular}} & \cellcolor[HTML]{EFEFEF}PaPr(z=0.3)  & \cellcolor[HTML]{EFEFEF}84.53     & \cellcolor[HTML]{EFEFEF}\textbf{160}  & \cellcolor[HTML]{EFEFEF}3x5                 \\
\bottomrule
\end{tabular}}
\end{minipage}
\hspace{1pt}
\begin{minipage}[b]{0.45\linewidth}\centering
\caption{Proposal ConvNet sweep in video recognition with ViT-B-MAE model as the baseline. PaPr achieves competitive performance with lighter proposal models for different \textit{$z$} values.}
\label{tab9}
\scalebox{0.74}{
\begin{tabular}{ccccc}
\toprule
\multicolumn{2}{c}{\textbf{Proposal}} & \multirow{2}{*}{\textbf{\begin{tabular}[c]{@{}c@{}}Keep\\ Ratio(z)\end{tabular}}} & \multirow{2}{*}{\textbf{\begin{tabular}[c]{@{}c@{}}Acc1\\ (\%)\end{tabular}}} & \multirow{2}{*}{\textbf{Views}} \\
\cmidrule(lr){1-2}
\textbf{Model}    & \textbf{GFLOPs}   &                                                                                   &                                    &                                 \\
\midrule
X3d-s~\cite{feichtenhofer2020x3d}             & 1.25              & 0.5                                                                               & 81.11                              & 3x5                             \\
X3d-m~\cite{feichtenhofer2020x3d}             & 2.45              & 0.5                                                                               & \textbf{81.19}                              & 3x5                             \\
\midrule
ResNet-50~\cite{feichtenhofer2019slowfast}         & 41.9              & 0.5                                                                               & 79.96                              & 3x5                             \\
ResNet-101~\cite{feichtenhofer2019slowfast}        & 85.67             & 0.5                                                                               & 79.97                              & 3x5                             \\
\midrule
X3d-s~\cite{feichtenhofer2020x3d}             & 1.25              & 0.4                                                                               & 80.86                              & 3x5                             \\
X3d-m~\cite{feichtenhofer2020x3d}             & 2.45              & 0.4                                                                               & 80.93                              & 3x5                             \\
\midrule
ResNet-50~\cite{feichtenhofer2019slowfast}         & 41.9              & 0.4                                                                               & 78.85                              & 3x5                             \\
ResNet-101~\cite{feichtenhofer2019slowfast}        & 85.67             & 0.4                                                                               & 79.32                              & 3x5                             \\
\bottomrule
\end{tabular}}
\end{minipage}
\vspace{-4mm}
\end{table}

\subsection{Experimental Setup}
\vspace{-2mm}

We experiment on image classification task on the ImageNet-1k benchmark dataset, following prior work~\cite{tome, adavit, ats, tokenfusion, xu2024gtp}. We use the MobileOne-s0~\cite{mobileone} model as the proposal model for all architectures, unless otherwise specified.  We report training-free results of PaPr, unless otherwise specified. For the performance metric, we report top-1 accuracy, GFLOPs, and the throughput (img/s). Throughput is measured on a single RTX-A5000 GPU with 24GB VRAM. We reproduced all baseline models under the same setup for a fair comparison.

\vspace{-3mm}
\subsection{Performance on Various Vision Transformers}
\vspace{-2mm}
We study performance of PaPr in diverse ViT architectures, along with various pre-training methods. 
We also present training results, along with training-free results to compare with existing methods.

\vspace{-3mm}

\subsubsection{Training-free method comparison.}
We consider two different pre-training methods for comparative analysis on various ViT architectures. We use supervised pretrained Augreg models~\cite{augreg}, and self-supervised pre-trained masked autoencoder (MAE) models~\cite{mae}, following prior work~\cite{tome}. 
We compare with three recent training-free patch reduction methods, such as ToMe~\cite{tome}, TokenFusion~\cite{tokenfusion}, and GTP-ViT~\cite{xu2024gtp}.
Additionally, we show the performance comparison on class-token free ViT models to highlight the architecture agnostic performance.
In general, PaPr achieves better accuracy, with lower computational costs compared to state-of-the-art (SOTA) methods.

\vspace{-2mm}
\paragraph{Augreg models:} For AugReg pre-training with ViT-Base, PaPr achieves 2.14\% higher accuracy over ToMe, with comparable FLOPs as shown in Tab.~\ref{tab1}. In addition, by combining PaPr with ToMe, we achieve additional 22.3\% FLOP reduction, while maintaining higher accuracy. We further study the compatibility of PaPr with existing patch reduction methods as ToMe. After the initial patch pruning with PaPr, we integrate ToMe only at the bottom-half layers due to its higher sensitivity in earlier layers. As shown in Fig.~\ref{fig:5}, integration of PaPr can directly boost ToMe performance thereby achieving Pareto optimality.

\vspace{-3mm}
\paragraph{MAE models:} 
MAE used self-supervised pre-training by reconstructing masked image patches to train large ViT models~\cite{mae}. 
Later, the model is fine-tuned with full resolution images without masking, which cannot exploit its latent ability to learn from fewer patches.
Interestingly, PaPr introduces masked inference, which is particularly suitable for MAE models. Hence, we observe significantly higher accuracy with PaPr compared to other training-free methods, \textit{e.g.}, with ViT-B, PaPr achieves 4.5\% higher accuracy over ToMe for similar FLOPs (see Tab.~\ref{tab2}).

\vspace{-3mm}
\paragraph{Class token-free models:}
Several existing patch reduction methods operate with the class token in ViTs to evaluate the patch relevance~\cite{ats, evit, adavit}. However, instead of class tokens, global pooling of patch tokens is used in many cases~\cite{hatamizadeh2023global, pan2021scalable}. Results presented in Tab.~\ref{tab4} demonstrate the superiority of PaPr over SOTA methods in similar models. 

\vspace{-4mm}
\subsubsection{Training-based methods comparison.}
We compare training performance of existing methods for the patch reduction on DeIT-s~\cite{deit} model in Tab.~\ref{tab3}. Prior pruning based methods cannot speed-up the training for learning the mask predictor~\cite{dynamicvit, spvit, Avit}. However, PaPr achieves competitive performance as prior methods, while achieving large speed-up as other token merging methods~\cite{tome, tokenfusion, xu2024gtp}.

\vspace{-3mm}
\subsection{Performance on Various Hierarchical Models}
\vspace{-2mm}
We study the performance on two variants of hierarchical models, such as pure convolutional models, and hybrid transformer models. We compare with training based DynamicCNN~\cite{dyncnn}, and DynamicSwin\cite{dyncnn} methods (trained for 120 epochs), whereas PaPr operates without training.

\vspace{-3mm}
\subsubsection{Analysis on convolutional models.}
We use the SOTA ConvNext~\cite{convnet2020} architecture for the analysis as shown in Tab.~\ref{tab5}. The training-free operation in PaPr makes it seamlessly usable with new model updates. We analyze ImageNet-1k and ImageNet-22k performance of same models. For the ConvNext-Base model, PaPr achieves 99.6\% accuracy of DynamicCNN with 7.1\% higher throughput. By simply using ImageNet-22k weights, PaPr can achieve additional 2\% accuracy improvement without training, while having the same computation cost.

\vspace{-3mm}
\subsubsection{Analysis on hybrid transformer models.}
We study the hierarchical Swin transformer models for patch reduction with PaPr, as given in Tab.~\ref{tab6}. PaPr achieves 98\% of DynamicSwin accuracy using similar FLOPs without re-training. Nevertheless, by leveraging bigger models, larger pre-training, and  higher resolution, PaPr can seamlessly adapt to achieve higher accuracy.

\vspace{-2mm}
\subsection{Robustness of PaPr across various ConvNet proposals}
PaPr can operate with ultra-lightweight ConvNets for generating robust proposals. We study different ConvNet architectures for proposal generation, as shown in Tab.~\ref{tab7}. Interestingly, PaPr shows small reductions of final accuracy (0.3\%), when using MobileOne-S0 based proposals compared to ResNet-152 ($42\times$ higher FLOPs) in ViT-B. When visualizing the PSM for different pruning ratio as in Fig.~\ref{fig:6}, we notice the similar PSMs irrespective of model size. Hence, PaPr can utilize ultra-lightweight ConvNets to mask patches without sacrificing accuracy.

\vspace{-3mm}
\subsection{Comparison with Class Activation Mappings (CAMs)}
Existing CAM methods mostly focus on explainability to highlight image regions responsible for final prediction. However, such objectives rely on final prediction performance, which can be much lower for light ConvNets. Moreover, for localization, many of these methods use gradients, and optimization methods  that limit batch processing. Nevertheless, we study the impact of such CAM methods (\cite{gradcam, basecam, scorecam}) in PaPr framework, in challenging scenarios where the baseline ViT-B has higher  prediction confidence (c) on target class, and the proposal MobileOne-S0 has lower confidence (see Fig.~\ref{fig:7}). In most cases, the baseline CAM method significantly lowers the final accuracy after patch pruning. In contrast, PaPr maintains robust confidence for its precise localization. Moreover, PaPr can enhance confidence of baseline models in several scenarios by suppressing the redundant patches, \textit{e.g.}, in Sample 5, the $c$ increased by 22\% with PaPr.

\begin{figure}[t]
\centering
\includegraphics[width= 0.93\textwidth]{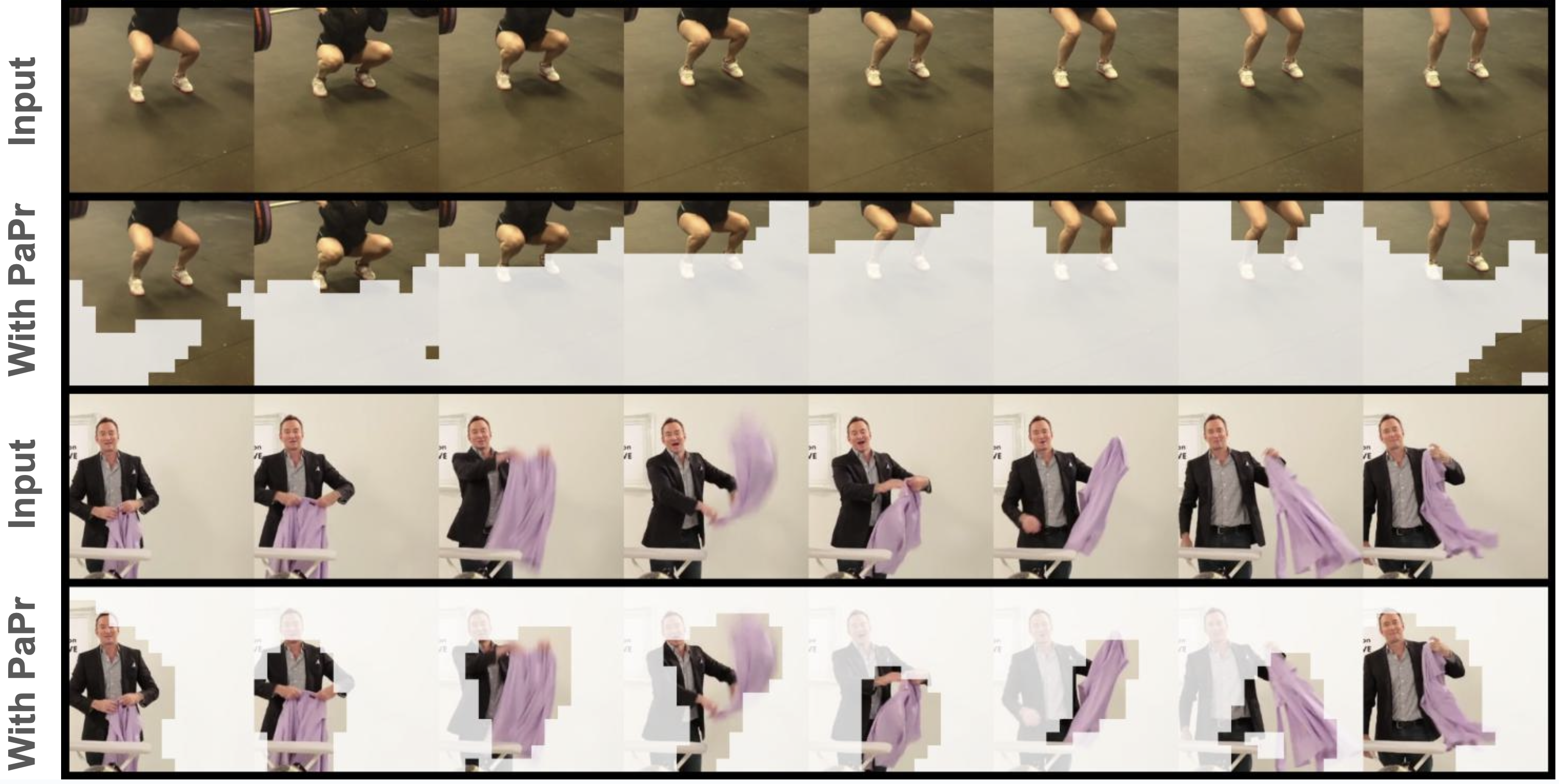}
\caption{Visualization of PaPr localization in videos. Video has inherent high sparsity. PaPr effectively localizes the discriminative regions for using holistic spatio-temporal understanding with small ConvNets. Thus, it significantly reduces the computational burden for larger models in downstream video recognition tasks.}
\label{fig:8}
\vspace{-5mm}
\end{figure}

\vspace{-3mm}
\section{Video Experiments}
\vspace{-2mm}
We study the training-free performance on Kinetics-400~\cite{kinetics} validation sets, as shown in Tab.~\ref{tab8}. We use SOTA
ATS~\cite{ats}, and ToMe~\cite{tome} methods for comparison. We use PaPr on SOTA ViT-MAE~\cite{videomae} models with lightweight X3d-s~\cite{feichtenhofer2020x3d} model for proposal generation. 
We follow the baseline~\cite{videomae} $3\times5$ view (3 spatial view and 5 temporal view) approach, and separate the views from model FLOP counts similar to other works as it can be arbitrarily chosen. Since ViT-MAE removes class tokens in fine-tuning, ATS~\cite{ats} cannot be adapted to these models. 
For ViT-L, PaPr achieves $3.7\times$ FLOPs reduction of the baseline model for a negligible 0.8\% accuracy drop.
We also study the impact of various proposal models on the final performance, as shown in Tab.~\ref{tab9}. We use SlowOnly-ResNet~\cite{feichtenhofer2019slowfast} and X3d~\cite{feichtenhofer2020x3d} models for comparison. Increasing model size for different pruning shows minimal impact on final performance.
Finally, we visualize the patch masking of PaPr in videos, as shown in Fig.~\ref{fig:8}. By removing the redundant details, PaPr significantly boosts the performance of bigger models for fine-grained predictions.

\label{experiments}

\vspace{-4mm}
\section{Conclusion and Future Works}
\label{conclusion}
\vspace{-2mm}

In this paper, we introduce a novel patch pruning method, namely PaPr, that can effectively speed-up \textit{off-the-shelf} pre-trained models inference without re-training. We propose a simple modification of ConvNets, that allows extracting a precise discriminative patch significance map (PSM) from ultra lightweight ConvNets with unparalleled speed. In PaPr, such PSMs are leveraged to directly reduce the redundant data regions in a single-step to guide larger model computations on most discriminative regions for achieving computational efficiency. Moreover, PaPr can be easily integrated with most existing patch reduction methods due to its simple structure. We validate the efficacy of PaPr through an extensive experimental study on diverse models with different architectures and pre-training schemes, as well as on different applications (image/video), which highlights the superiority of PaPr over existing patch reduction methods. 

Although we studied the use of PaPr  on discriminative tasks only, it has great potential for dense prediction tasks, which we leave for future study.


%
%
\bibliographystyle{splncs04}
\bibliography{main}
\appendix
\section{Appendix}
We provide additional qualitative analysis to illustrate the effectiveness of PaPr in practice (Sec.~\ref{sa}--Sec.~\ref{sc}). Additionally, we provide simple PyTorch pseudo code for PaPr implementation (Sec.~\ref{sd}).

\section{Robustness of PaPr with Various ConvNet Proposals}
\label{sa}
\begin{figure}[t]
\centering
\includegraphics[width= 1.0\textwidth]{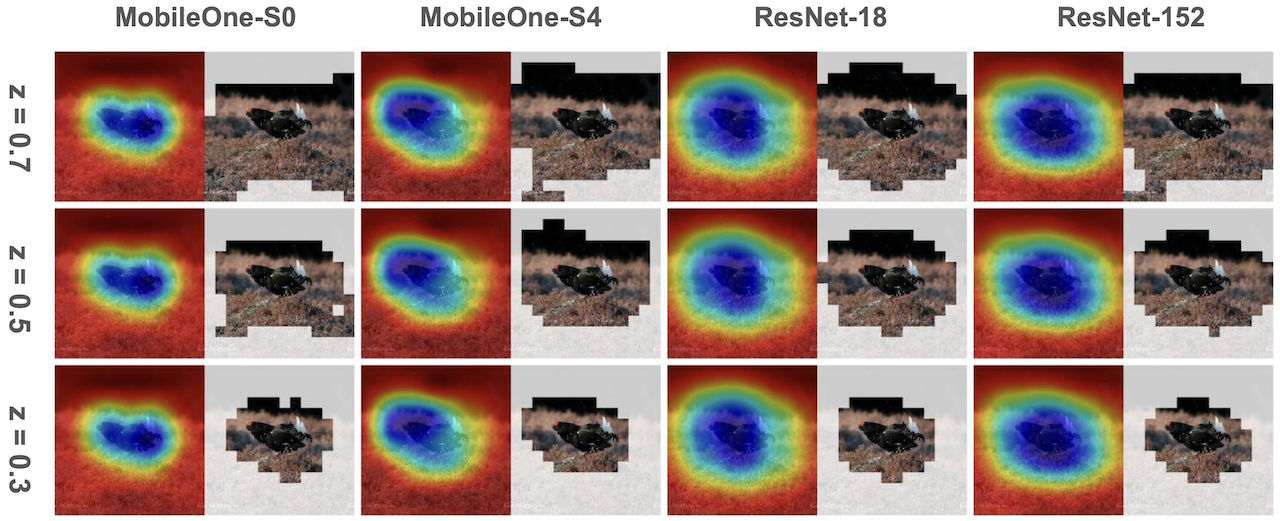}
\caption{More visualizations of patch significance map (PSM) and patch masks with various proposal models for different keeping ratio (\textit{z}).}
\label{fig:s1}
\end{figure}

\begin{figure}[t]
\centering
\includegraphics[width= 1.0\textwidth]{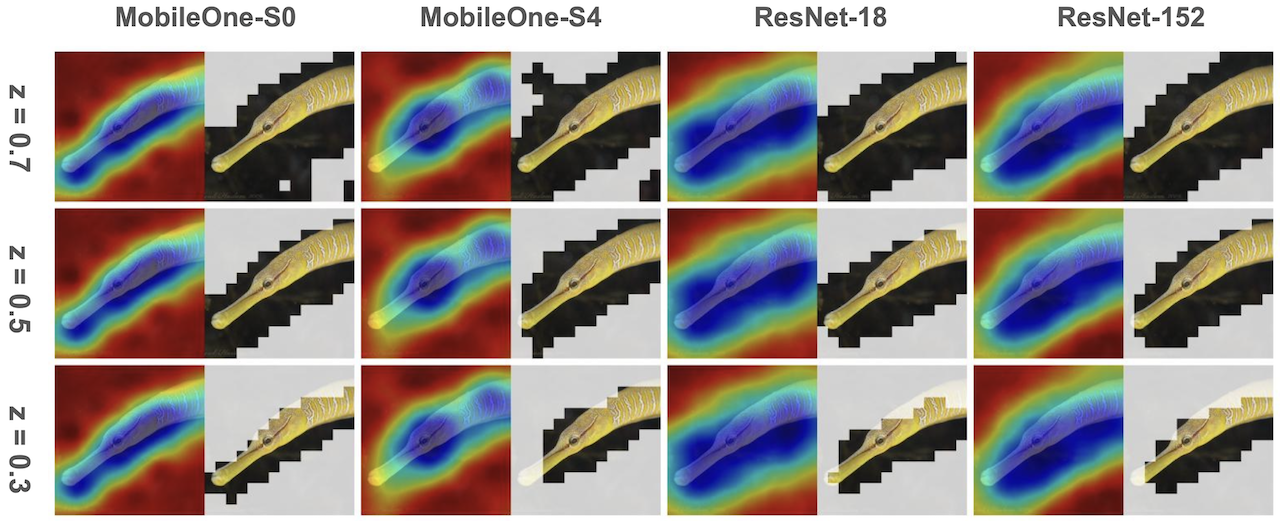}
\caption{More visualizations of patch significance map (PSM) and patch masks with various proposal models for different keeping ratio (\textit{z}).}
\label{fig:s2}
\end{figure}

\begin{figure}[t]
\centering
\includegraphics[width= 1.0\textwidth]{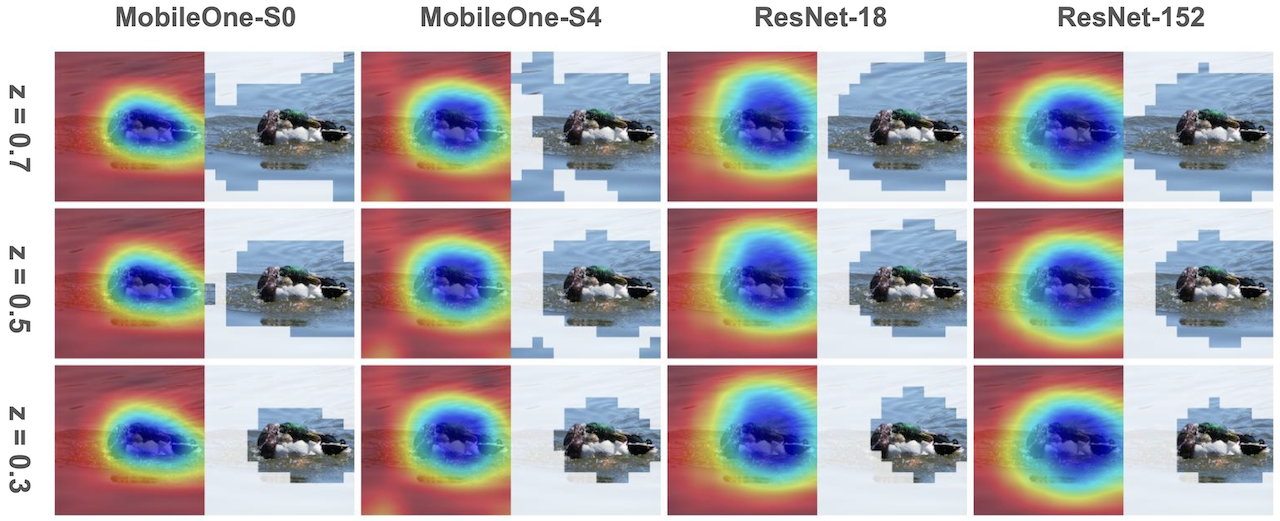}
\caption{More visualizations of patch significance map (PSM) and patch masks with various proposal models for different keeping ratio (\textit{z}).}
\label{fig:s3}
\end{figure}

We study the patch significance map (PSM), and patch masks generated by different ConvNet proposal networks. To make PaPr more computationally efficient and accurate, we need precise mask of discriminative regions irrespective of the size and top-1 accuracy of the proposal network. In Fig.~\ref{fig:s1}, Fig.~\ref{fig:s2}, and Fig.~\ref{fig:s3}, we demonstrate more visualizations of generated PSMs and patch masks with different keeping ratios for various ConvNets. Despite little variations in PSMs for different ConvNets, top $z\%$ patch mask almost remains identical focusing the key image patches. With $z=0.5$, all proposal models visually represent robust performance to keep the key patches representing target objects. With much lower keeping ratio as $z=0.3$, few parts of the object in Fig.~\ref{fig:s2} are masked, however, in Fig.~\ref{fig:s1} and Fig.~\ref{fig:s3}, most key object patches are visible even with such high masking ratio. This highlights that PaPr can operate with extremely lightweight ConvNet (MobileOne-S0 has $42\times$ smaller FLOPs than ResNet-152) for precise key discriminative patch localization, that makes it particularly suitable for larger models to prune redundant image patches.

\section{Robustness of PaPr over CAM Methods}
\label{sb}
Class activation mapping (CAM) methods mostly focus on highlighting key image patches responsible for the target class prediction to make the decision more interpretable~\cite{basecam, scorecam, gradcam, gradcam++, bettercam}. Such CAM methods have two major limitations to be useful for patch masking: (1) These methods cannot leverage batch processing for separately tracing the sample activation for each prediction. Moreover, many of these methods rely on gradient modulation~\cite{gradcam, gradcam++}, or complex activation decomposition~\cite{scorecam, eigencam} that practically make them infeasible to speed-up large \textit{off-the-shelf} models.
(2) Since these methods heavily rely on activation weights of the final prediction (usually, in the final FC layer), it makes them particularly problematic for smaller models with significantly lower top-1 accuracy. Therefore, rather than highlighting the class regions based on final prediction, PaPr attempts to localize the most discriminative patch regions irrespective of its class, that makes it suitable for ultra-lightweight proposal ConvNets unaffected by its size or final top-1 accuracy. 

\begin{figure}[t]
\centering
\includegraphics[width= 1.0\textwidth]{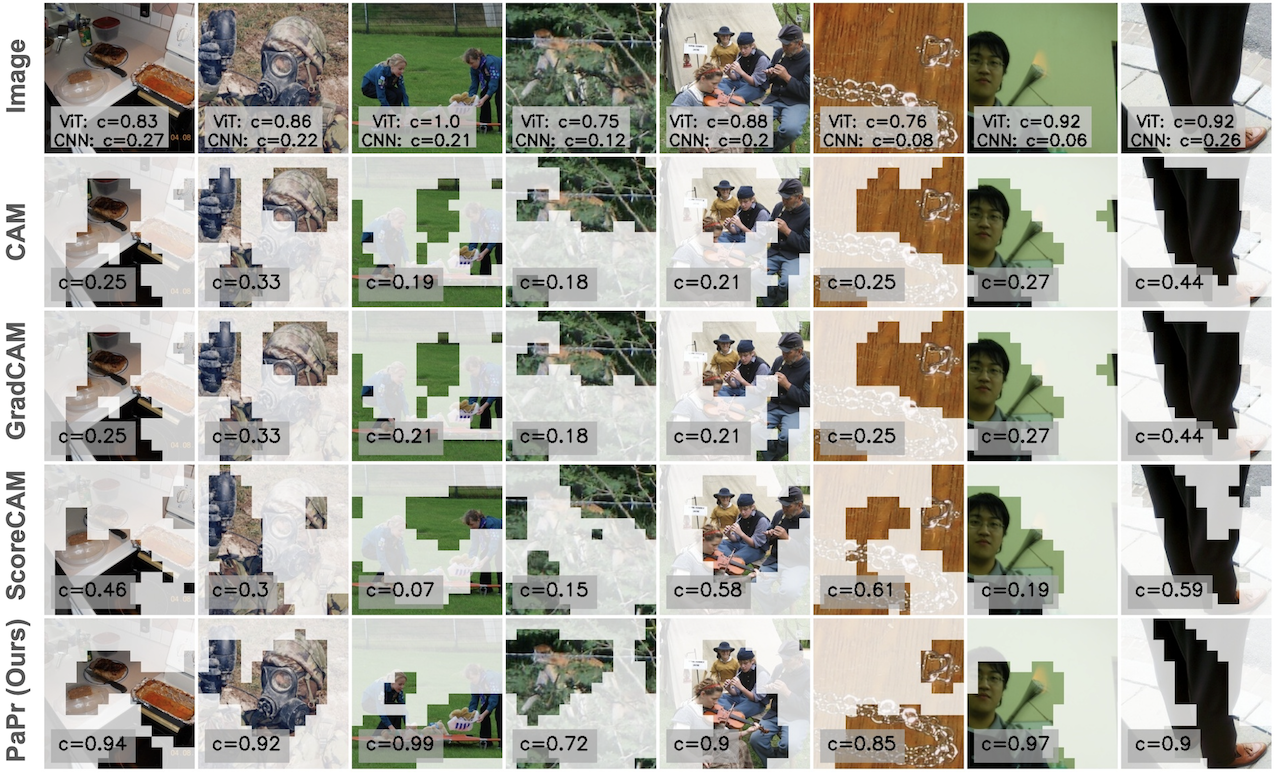}
\caption{More visualizations on robustness of PaPr compared to CAM based methods.}
\label{fig:s4}
\vspace{-5mm}
\end{figure}

\begin{figure}[t]
\centering
\includegraphics[width= 1.0\textwidth]{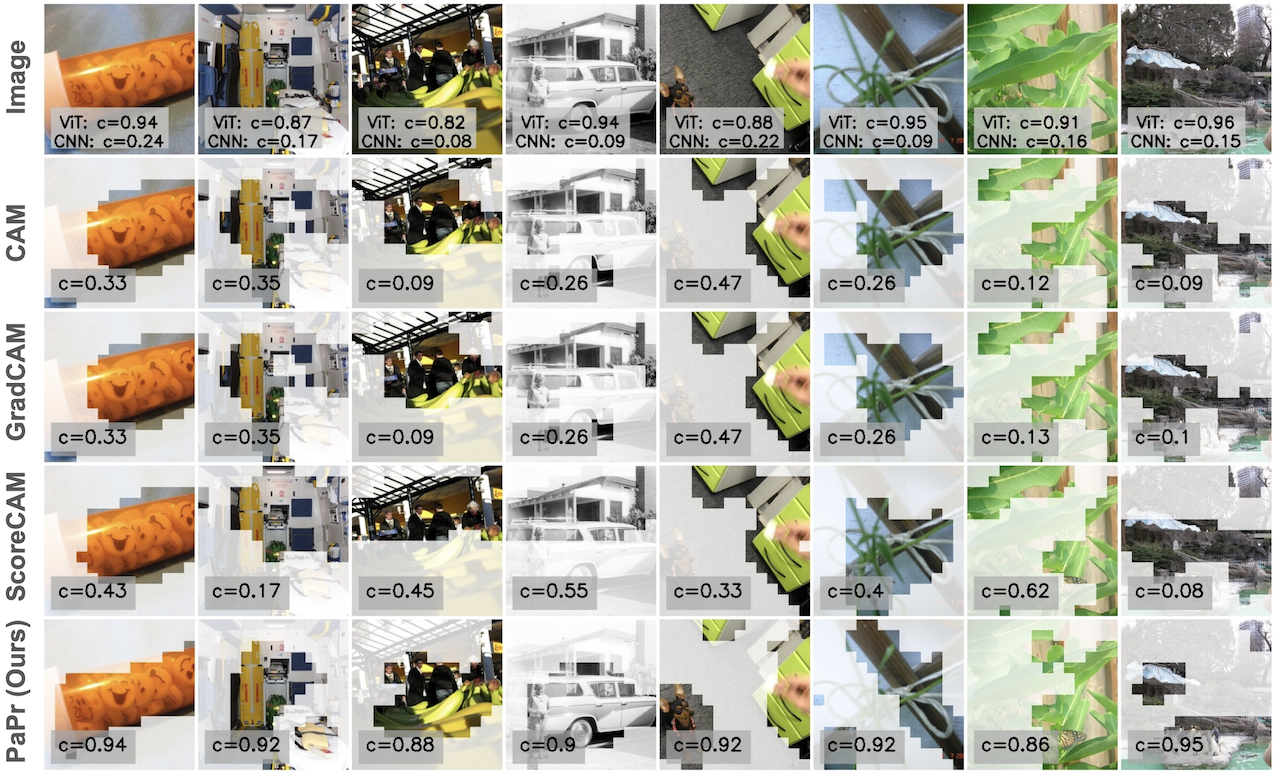}
\caption{More visualizations on robustness of PaPr compared to CAM based methods.}
\label{fig:s5}
\vspace{-5mm}
\end{figure}

\begin{figure}[t]
\centering
\includegraphics[width= 1.0\textwidth]{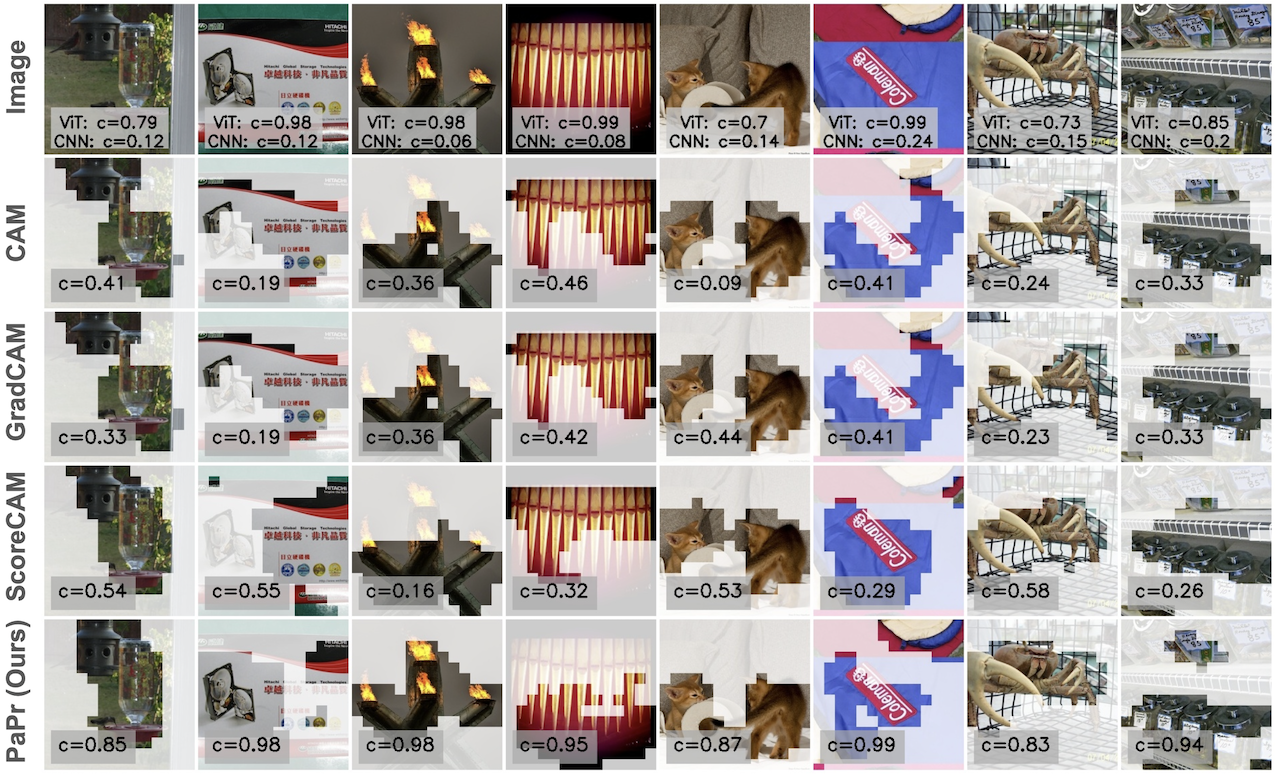}
\caption{More visualizations on robustness of PaPr compared to CAM based methods.}
\label{fig:s6}
\vspace{-5mm}
\end{figure}

\begin{figure}[t]
\centering
\includegraphics[width= 1.0\textwidth]{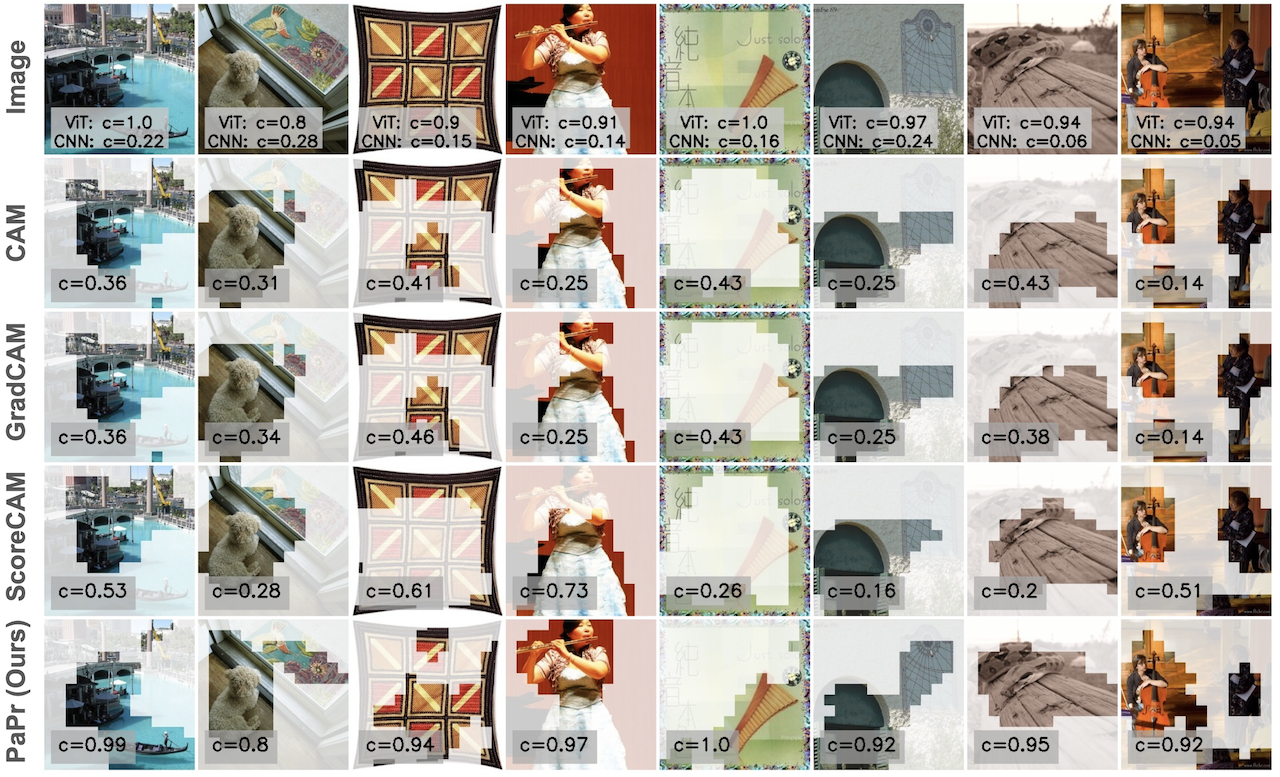}
\caption{More visualizations on robustness of PaPr compared to CAM based methods.}
\label{fig:s7}
\vspace{-5mm}
\end{figure}

\begin{figure}[t]
\centering
\includegraphics[width= 1.0\textwidth]{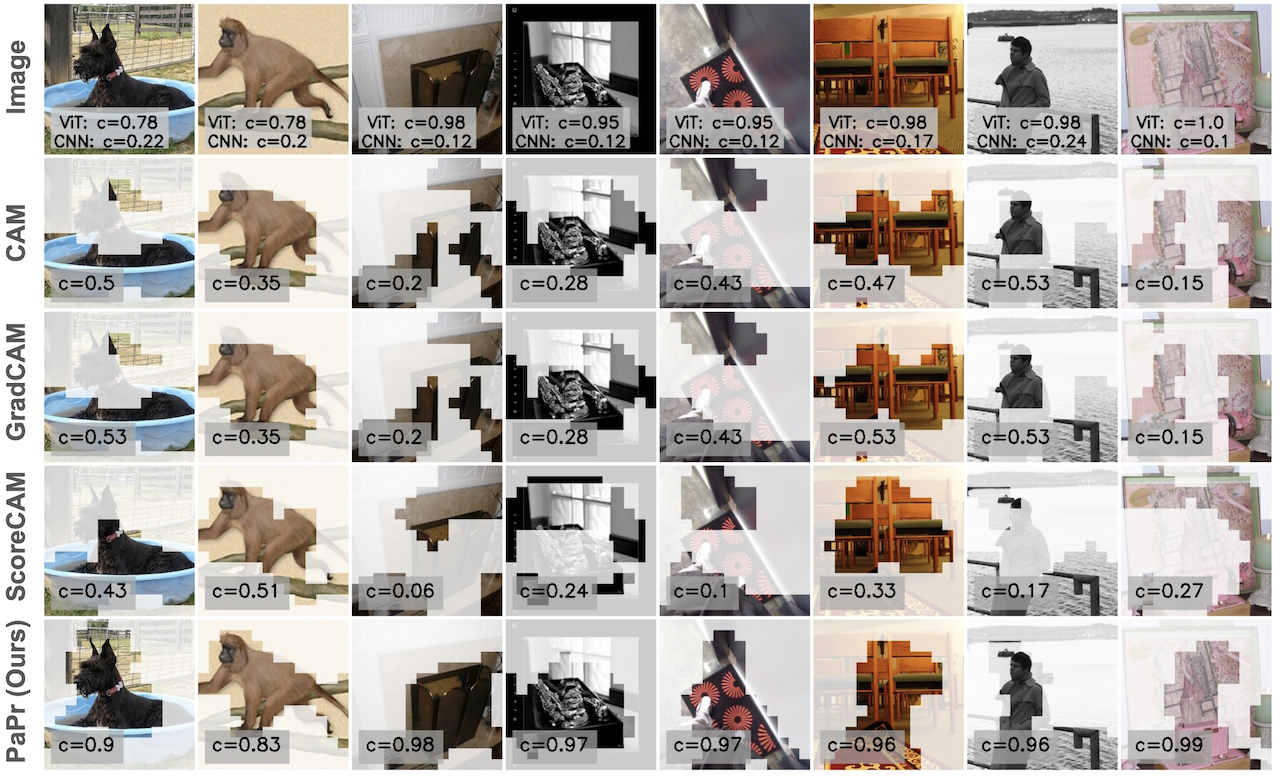}
\caption{More visualizations on robustness of PaPr compared to CAM based methods.}
\label{fig:s8}
\vspace{-5mm}
\end{figure}

\begin{figure}[t]
\centering
\includegraphics[width= 1.0\textwidth]{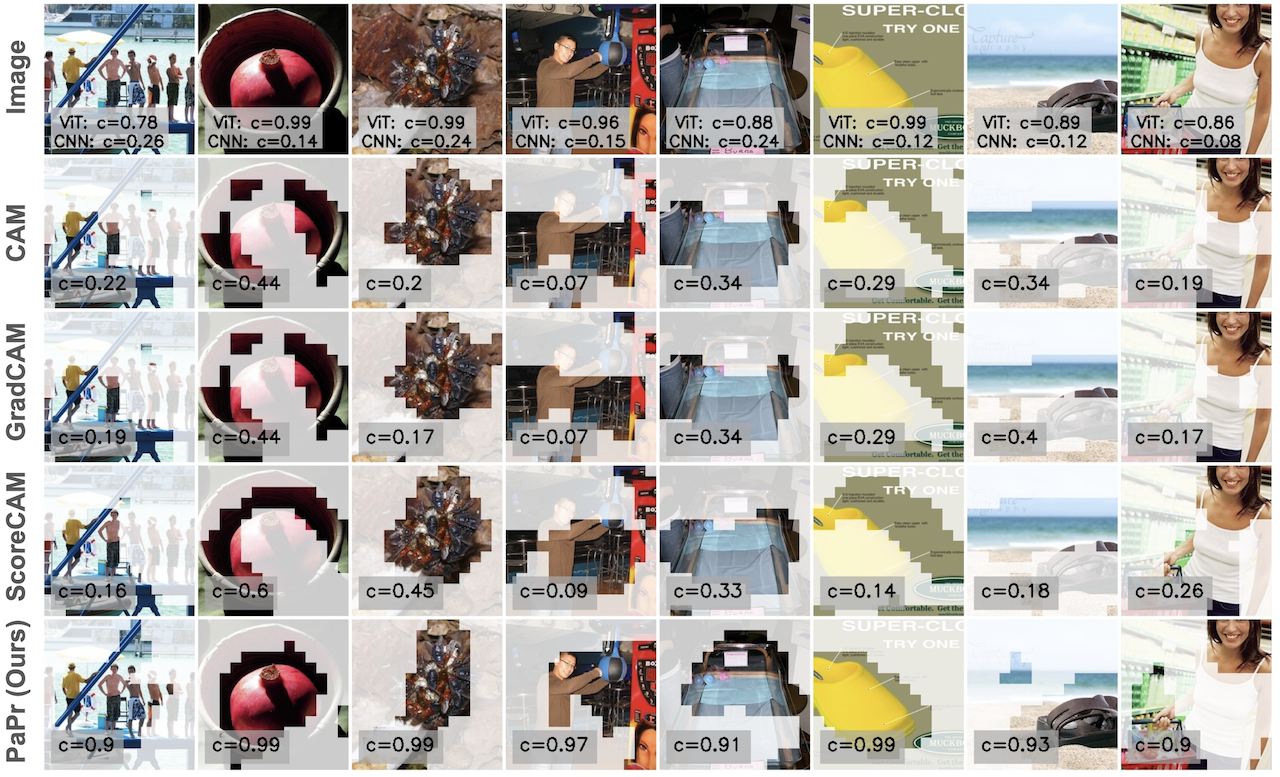}
\caption{More visualizations on robustness of PaPr compared to CAM based methods.}
\label{fig:s9}
\vspace{-5mm}
\end{figure}

We provide extensive qualitative comparisons of various CAM methods with PaPr in Fig.~\ref{fig:s4}--Fig.~\ref{fig:s9}. To focus on more challenging samples, we mainly present the results on images, where the proposal ConvNet has significantly lower confidence on the target class while the larger ViT has significantly higher confidence. We denote the final confidence $c$ on the target class in each sample. We use ViT-Base-16 model as the baseline, and lightweight MobileOne-S0 as the proposal model. We analyze whether the application of patch masking with light ConvNet (MobileOne-S0 in our example) has significant impact on the target class prediction of the large model (ViT-Base-16 in our example). We use keeping ratio of $z=0.4$ to keep top-40\% discriminative patches in each method. We highlight several key findings from these qualitative analysis: (1) PaPr can maintain the prediction confidence of large ViTs with significantly smaller amount of patches, whereas other CAM based methods face significant reduction of confidence, mostly in challenging cases. (2) Notably, we observe the increase of prediction confidence in several cases with reduced patches. We hypothesize that, such patch masking greatly reduces the complexity of the image by masking the backgrounds that results in increase of confidence. (3) PaPr performs significantly better particularly in cases where the ConvNet has extremely low confidence, whereas other CAM methods struggle in such scenarios. These demonstrate PaPr's ability to precisely locate the key discriminative patches in challenging scenarios without hurting the large model's performance.
\vspace{-0.2cm}
\section{Qualitative Analysis on Patch Pruning in Videos}
\label{sc}
\begin{figure}[t]
\centering
\includegraphics[width= 1.0\textwidth]{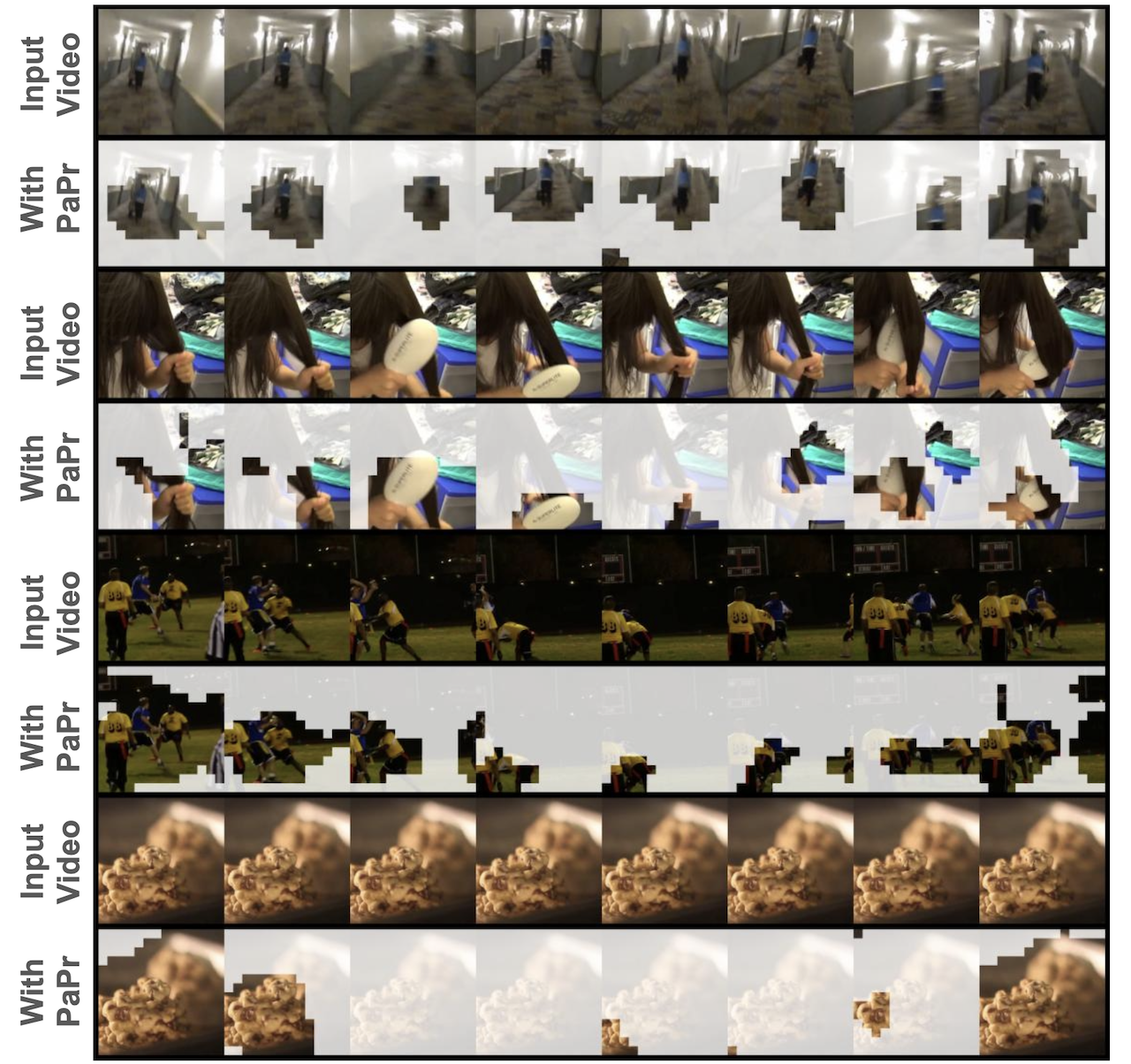}
\caption{More visualizations of spatio-temporal patch masking in videos with PaPr for keeping ratio $z=0.3$. X3d-s~\cite{feichtenhofer2020x3d} based ConvNet is used for visualization.}
\label{fig:s10}
\vspace{-5mm}
\end{figure}

\begin{figure}[t]
\centering
\includegraphics[width= 1.0\textwidth]{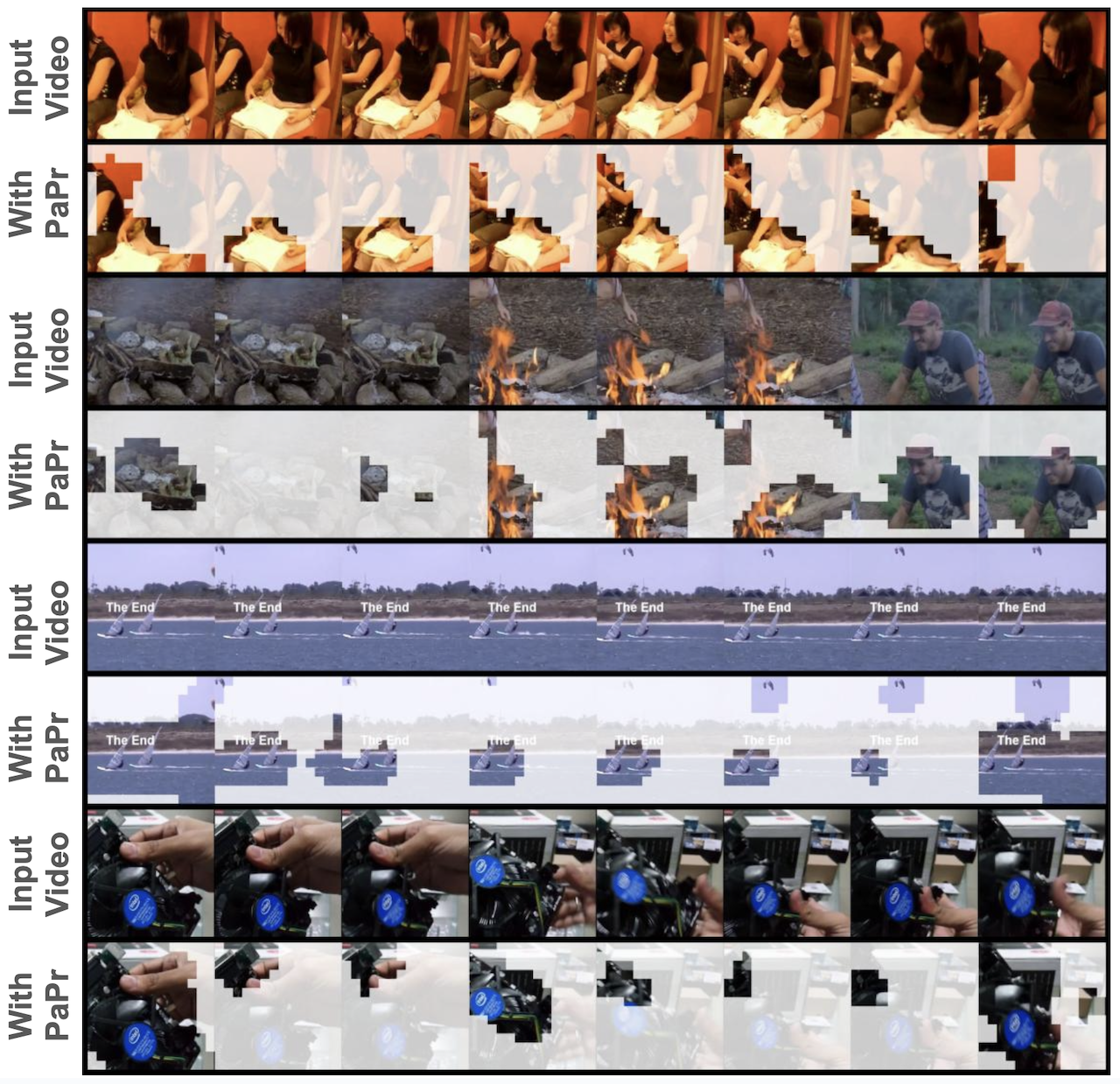}
\caption{More visualizations of spatio-temporal patch masking in videos with PaPr for keeping ratio $z=0.3$. X3d-s~\cite{feichtenhofer2020x3d} based ConvNet is used for visualization.}
\label{fig:s11}
\vspace{-5mm}
\end{figure}

In general, video contains large information redundancies, particularly for the video recognition task, that makes such applications computationally burdensome for larger models. However, to locate key discriminative patch regions in videos, spatio-temporal perception of the whole video is required. Interestingly, we can integrate PaPr with light ConvNets for background patch masking with spatio-temporal reasoning to speed-up larger models. In Fig.~\ref{fig:s10} and Fig.~\ref{fig:s11}, we provide additional visualizations on spatio-temporal patch masking in videos with PaPr. We use lightweight X3d-s~\cite{feichtenhofer2020x3d} model with low patch keeping ratio of $z=0.3$ for visualization. We highlight the major observations as follows: (1) PaPr can track patches representing the target object across complex backgrounds. (2) In slow moving videos with similar backgrounds, PaPr reduces the data redundancy by suppressing similar frames. Usually, the starting and ending frames are observed with higher priority, while similar intermediate frames are heavily masked.
(3) PaPr can precisely isolate few frames representing the main object regions across other redundant frames, that requires holistic understanding of the whole video. These results demonstrate that, PaPr can be very effective in suppressing redundant spatio-temporal patches to significantly reduce computational burden of large \textit{off-the-shelf} models in video recognition.

\pagebreak
\section{PyTorch Implementation}
\label{sd}
We provide pseudo code implementation of PaPr in PyTorch \cite{pytorch} on class-token based vision transformer~\cite{vit}. In particular, we apply PaPr to prune redundant patch tokens in ViTs after the initial extraction of patch tokens with the integration of class token and position embedding.  Starting from the ViT tokens, and final convolutional feature maps extracted from the proposal ConvNet, the following code snippet can prune redundant patch tokens with target keeping ratio (z). We note that, PaPr can operate with class-token free ViTs~\cite{vit}, hierarchical transformers~\cite{swin}, pure ConvNets~\cite{convnet2020}, and video transformers~\cite{videomae}. 

\begin{python}
def apply_papr(x: torch.tensor, f: torch.tensor, z: float) -> torch.tensor:
"""
    x: input ViT tokens of size (batch, N, c)
    f: proposal ConvNet features of size (batch, K, h, w)
    z: keeping ratio for tokens
"""  
    b, n, c = x.shape
    h1 = w1 = numpy.sqrt(n-1) # spatial resolution of tokens
    nt = int(n*z) # total remaining tokens after pruning

    # extract discriminative feature map from proposal features
    Fd = f.mean(dim=1) # size (batch, h, w)
    
    # upsampling F to match patch token spatial resolution in x
    # it generates Patch Significance Map (P)
    import torch.nn.functional as F
    P = F.interpolate(Fd, size=(h1, w1), mode="bicubic") 
    P = P.view(b, -1) # reshaping for pruning mask extraction

    # extracting indices of the most significant patches
    patch_indices = P.argsort(dim=1, descending=True)[:, :nt] 

    patch_indices += 1 # adjusting indices for class tokens

    # preparing class indices for each sample
    class_indices = torch.zeros(b, 1).to(patch_indices.device)

    # Patch mask is obtained combining class and patch indices
    M = torch.cat([class_indices, patch_indices], dim=1)
    
    # extracting tokens based on patch mask
    x = x.gather(dim=1, index=M.unsqueeze(-1).expand(b, -1, c))

    # pruned x tensor size (batch, nt, c)
    return x

\end{python}


\end{document}